\documentclass[letterpaper]{article} 
\usepackage{aaai24}  
\usepackage{times}  
\usepackage{helvet}  
\usepackage{courier}  
\usepackage[hyphens]{url}  
\usepackage{graphicx} 
\usepackage{natbib}  
\usepackage{caption} 
\usepackage{algorithm}
\usepackage{algorithmic}
\usepackage{algorithm, soul, kotex}
\usepackage{amsmath,amssymb}
\usepackage{dsfont}
\usepackage{booktabs,tabularx}
\usepackage{color, colortbl}
\definecolor{backcolour}{RGB}{230, 230, 230}
\usepackage{appendix}
\usepackage{comment}
\usepackage{xcolor}
\definecolor{myred}{rgb}{1,0,0}
\let\oldref\ref
\renewcommand{\ref}[1]{\textcolor{myred}{\oldref{#1}}}

\def\1{\mathds{1}}
\usepackage{newfloat}
\usepackage{listings}
\DeclareCaptionStyle{ruled}{labelfont=normalfont,labelsep=colon,strut=off} 
\floatstyle{ruled}
\newfloat{listing}{tb}{lst}{}
\floatname{listing}{Listing}
\newcounter{alphasect}
\def\alphainsection{0}

\let\oldsection=\section
\def\section{%
  \ifnum\alphainsection=1%
    \addtocounter{alphasect}{1}
  \fi%
\oldsection}%

\renewcommand\thesection{%
  \ifnum\alphainsection=1%
    \Alph{alphasect}
  \else%
    \arabic{section}
  \fi%
}%

\title{All but One: Surgical Concept Erasing with Model Preservation in \\Text-to-Image Diffusion Models}
\author{
    Seunghoo Hong\textsuperscript{\rm 1}\equalcontrib,
    Juhun Lee\textsuperscript{\rm 1}\equalcontrib,
    Simon S. Woo\textsuperscript{\rm 1}
}
\affiliations{
    \textsuperscript{\rm 1}Department of Artificial Intelligence, Sungkyunkwan University, S. Korea\\
    hoo0681@g.skku.edu, josejhlee@g.skku.edu, swoo@g.skku.edu

}

\usepackage{bibentry}

\begin{document}

\twocolumn[{
\renewcommand\twocolumn[1][]{#1}%
\maketitle

\begin{center}
\vspace{-25pt}
    \centering
    \captionsetup{type=figure}
    \includegraphics[width=0.85\linewidth]{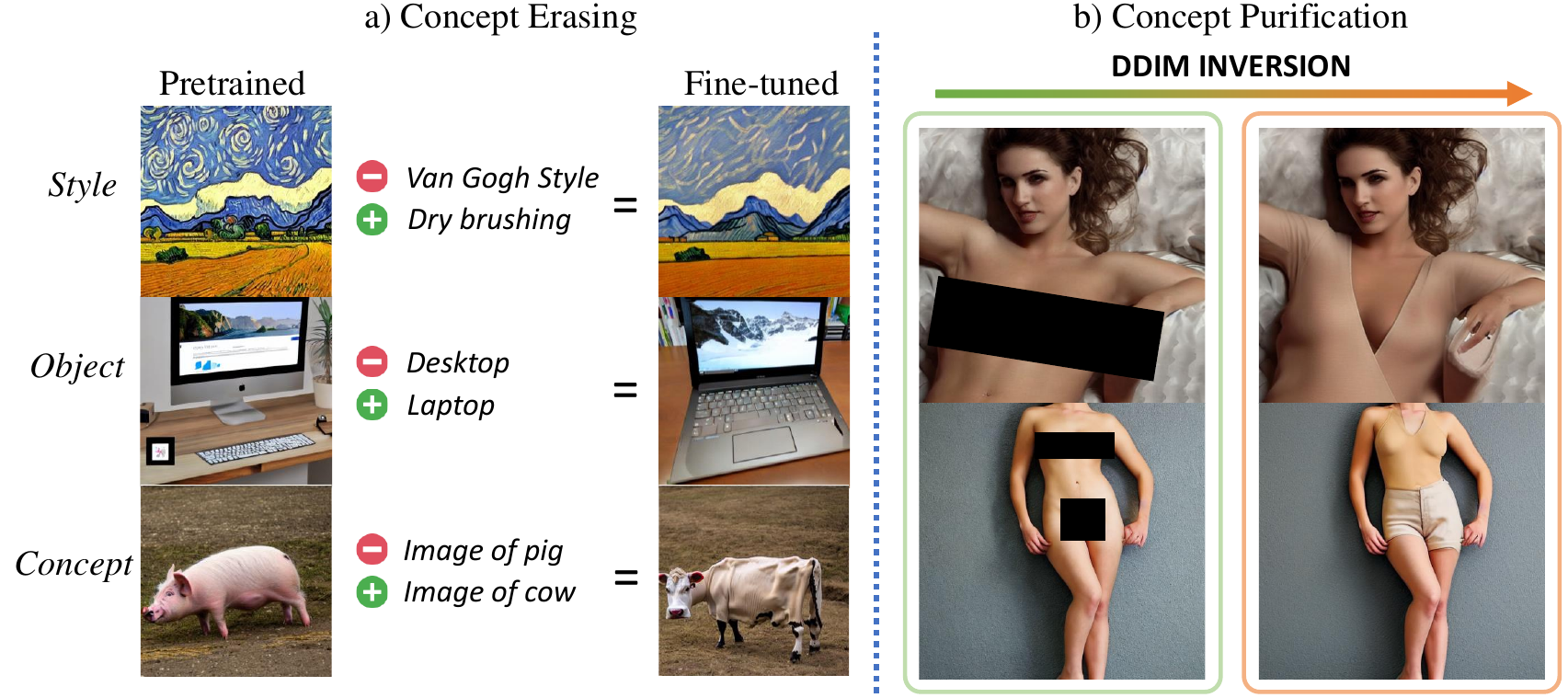}
    \captionof{figure}{Our method erases a concept while leaving the model intact. To enable more controllability in the erasing procedure, the user may introduce alternative concepts. Also, it fixes many current issues such as spatial inconsistency, model degradation, and training inefficiency that have been the problems in many previous approaches. Interestingly, our model can even “purify” and censor concepts with DDIM inversion, which is currently hardly reproducible through other methods to practically erase nudity from real-world data.}

    \label{fig:main_figure}
\end{center}%
}]


\begin{abstract}
Text-to-Image models such as Stable Diffusion have shown impressive image generation synthesis, thanks to the utilization of large-scale datasets. However, these datasets may contain sexually explicit, copyrighted, or undesirable content, which allows the model to directly generate them. Given that retraining these large models on individual concept deletion requests is infeasible, fine-tuning algorithms have been developed to tackle concept erasing in diffusion models. While these algorithms yield good concept erasure, they all present one of the following issues: 1) the corrupted feature space yields synthesis of disintegrated objects, 2) the initially synthesized content undergoes a divergence in both spatial structure and semantics in the generated images, and 3) sub-optimal training updates heighten the model's susceptibility to utility harm. These issues severely degrade the original utility of generative models. In this work, we present a new approach that solves all of these challenges. We take inspiration from the concept of classifier guidance and propose a surgical update on the classifier guidance term while constraining the drift of the unconditional score term. Furthermore, our algorithm empowers the user to select an alternative to the erasing concept, allowing for more controllability. Our experimental results show that our algorithm not only erases the target concept effectively but also preserves the model’s generation capability. 
\end{abstract}
\section{Introduction}

Recently, large-scale text-to-image models have demonstrated a remarkable ability to synthesize photo-realistic images \cite{rombach2022high,saharia2022photorealistic,ramesh2021zero}. This rise in generative models was elicited by the joint advancement of algorithms, computing resources, and the curation of large-scale datasets such as  LAION~\cite{schuhmann2022laion}. While these datasets offer rich features for training large-scale models~\cite{brown2020language,dosovitskiy2020image}, many of them are curated with web-scraped material and, thus, lack the necessary preprocessing regarding safety,  privacy, and bias \cite{mehrabi2021survey}. Moreover, such datasets often contain sexually explicit content, copyrighted material, and personal images. Training generative models using these sensitive data means that the model’s generative capability is derived from these same images, and the model is capable of generating such inappropriate content partially or entirely. 

To make things worse, due to the stochastic property and the capability to model complex distributions, there is always a non-zero likelihood that the generated image will contain unsafe content even when conditioned on any unrelated text token. This limits the usability of these generative models in public settings. To alleviate this problem, researchers have inserted an NSFW safe-checking neural network \cite{von-platen-etal-2022-diffusers}. Still, alongside a high false positive rate, their near-unpredictable masking rate of images limits their applicational range, especially when the application relies on a continuous stream of data~\cite{rando2022red}. Given these complications, both computation-wise and performance-wise, of retraining these “foundational” models with a heavily curated dataset, researchers have proposed to directly fine-tune foundation models such as Stable Diffusion to erase target concepts~\cite{gandikota2023erasing,kumari2023ablating,zhang2023forget}. 

While such fine-tuning algorithms for concept ablation are efficient in erasing itself, they significantly sacrifice much of the original generative power of the model to do so. This is far from the original motivation of this line of research. Through a closer examination of the utility aspect of these models, we identify three issues with the current fine-tuning algorithms: 1) Due to the corruption in the feature space of the model, generated images prompted with any arbitrary concepts become unrecognizable or very different from their original concept (see Fig.~\ref{fig:object_desintegration}),
2) Generative models such as diffusion models rely on random seeds to output images. As a consequence of fine-tuning the model, the spatial structure and the semantics of the image from the same random seed change. If we regard the model before erasing as the oracle, then any unintended deviation in the output image is not aligned with the ultimate utility of ablating concepts in the model, and 3) despite the model displaying adequate erasing capabilities, certain methods demand a high number of iterations, thereby subjecting the model to increased utility harm. Recent algorithms~\cite{gandikota2023erasing,kim2023towards} for erasing recommend around 1,000 update steps, which increases the exposure to the issues mentioned above.

In this paper, we aim to address all of the aforementioned challenges. Our main motivation comes from the hypothesis that the task of erasing a concept while preserving the rest requires a \textit{surgical intervention}, where we modify the concept of interest no more than needed. To achieve this, we first inherit from the idea of \textit{classifier guidance}\cite{ho2022classifier} to decompose the intermediate latent into the unconditional score and the guidance score term and solely apply updates to the latter term. In this region of update, we modify the target concept by introducing supervised and unsupervised erasing guidance, which shows that updating the guidance score is agnostic to the method of erasing guidance supervision method. Moreover, deriving from the Lagrangian Multiplier method, we introduce a regularization on the unconditional score term so that it does not interfere with the update of the guidance score distribution.

Our main contributions are summarized as follows:
\begin{itemize}

    \item We examine the possible societal and harmful effects of the latest generative models and approaches to mitigate issues via concept erasing, especially focusing on sexually explicit content. We identify that current SoTA algorithms do not consider model utility enough when erasing a concept, and most of them fall short of being used for practice. 

    \item We formulate a fine-tuning algorithm that modifies the core of the target concept while keeping the model intact. Our approach naturally gives rise to a regularization term, where we effectively and safely control the trade-off between erasing strength and model preservation.

    \item Through extensive experiments, we demonstrate that our surgical approach improves on spatial and semantic consistency, and training efficiency over the current baselines with FID, KID, CLIP, and SSIM scores.
    
\end{itemize}
\section{Background}

We first describe the essential components used in this line of research before explaining relevant research works.
\subsection{Diffusion Models}
Diffusion models are a class of generative models that learn to reverse the Markov chain diffusion process. Let $x_0$ represent true data observations and $x_t$ represent intermediate noised data, when $t = T$, corresponding observations $x_T$ are noised to become Gaussian noise. More precisely, diffusion process~\cite{ho2020denoising,song2020denoising} is defined as
\begin{equation}
\begin{aligned}
q\left(x_t \mid x_{t-1}\right) & := \mathcal{N}\left(x_t ; \sqrt{\alpha_t} x_{t-1},\left(1-\alpha_t\right) \mathbf{I}\right) \\
q(x_T|x_0)&\approx\mathcal{N}(x_T;\mathbf{0},\mathbf{I}), \\
\end{aligned}
\end{equation}
where $\alpha_t$ is a fixed~\cite{ho2020denoising} or learnable schedule~\cite{sohl2015deep}. According to Bayes' rule, we can obtain the reverse diffusion, which can be interpreted as an interpolation between $x_t$ and $x_0$. Then, we can learn to predict this distribution by matching it with a parameterized network and minimizing the KL divergence of the two distributions. The divergence of two Gaussian distributions can be formulated as the mean square error loss. In practice, we reparameterize $x_t$ so that we predict the epsilon $\epsilon_t$~\cite{ho2020denoising} that was used to generate $x_t$ as follows:
\begin{equation}
\mathcal{L}_\text{diffusion}=\mathbb{E}_{x_t, t, \epsilon \sim \mathcal{N}(0,1)}\left[\left\|\epsilon-\epsilon_\theta\left(x_t, t\right)\right\|_2^2\right]
\end{equation}

\subsection{Text-to-Image Diffusion Models}

By diffusing in the latent space of powerful VAEs~\cite{oord2017neural,razavi2019generating} and conditioning these models with text embeddings~\cite{ramesh2021zero}, they take the form of Latent Diffusion Models (LDM)~\cite{rombach2022high,saharia2022photorealistic} or commonly known as “text-to-image diffusion models”. With the addition of these two components, the loss can be formulated as follows:
\begin{equation}
\mathcal{L}_\text{LDM}=\mathbb{E}_{z_t \in \mathcal{E}(x), t, c, \epsilon \sim \mathcal{N}(0,1)}\left[\left\|\epsilon-\epsilon_\theta\left(z_t, c, t\right)\right\|_2^2\right],
\end{equation}
where $z_t$ is the noised latent embedding of image $x$ through a VAE, and $c$ is the text embedding encoded by text encoders such as CLIP~\cite{radford2021learning}. 

\subsection{Classifier guidance and Classifier-free guidance}
It is well known that score $-\sigma_t \nabla_{\boldsymbol{z_t}
}\log p(z_t)$ and epsilon $\boldsymbol{\epsilon}_\theta(z_t)$ are equivalent. Then, given that $p_{\theta}(z_{t}|c)p_{\theta}(c|z_{t})^{\gamma} \propto p_{\theta}(z_{t})p_{\theta}(c|z_{t})^{\gamma+1}$~\cite{ho2022classifier,song2020score,dhariwal2021diffusion}, we can formulate classifier guidance as follows:
\begin{equation}
\begin{aligned}
\tilde{\boldsymbol{\epsilon}}_\theta\left(\mathbf{z}_t|\mathbf{c}\right)&=\boldsymbol{\epsilon}_\theta\left(\mathbf{z}_t\right)-(\gamma+1) \sigma_t \nabla_{\mathbf{z}_t} \log p_\theta\left(\mathbf{c} \mid \mathbf{z}_t\right)\\  &\approx-\sigma_t \nabla_{\mathbf{z}_t}\left[\log p\left(\mathbf{z}_t\right)+(\gamma+1) \log p_\theta\left(\mathbf{c} \mid \mathbf{z}_t\right)\right] \\
& =-\sigma_t \nabla_{\mathbf{z}_t}\left[\log p\left(\mathbf{z}_t \mid \mathbf{c}\right)+\gamma \log p_\theta\left(\mathbf{c} \mid \mathbf{z}_t\right)\right].
\end{aligned}
\end{equation}
With classifier-free guidance (CFG)~\cite{ho2022classifier}, one can obtain $\nabla_{\mathbf{z}_t}\log p(\mathbf{c}\mid\mathbf{z}_t)$ by composing the scores $\boldsymbol{\epsilon}_\theta(z_t)$ and $\boldsymbol{\epsilon}_\theta(z_t,c)$ as follows
\begin{equation}
\nabla_{\mathbf{z}_t}\log p(\mathbf{c}\mid\mathbf{z}_t) = -\frac{1}{\sigma_t}[\boldsymbol{\epsilon}_\theta(z_t,c) - \boldsymbol{\epsilon}_\theta(z_t)].
\end{equation} 
Ultimately, we can sample an epsilon with guidance scale, $\gamma$, as follows: 
\begin{equation}
\tilde{\boldsymbol{\epsilon}}_\theta(z_t,c) = (1+\gamma)\boldsymbol{\epsilon}_\theta(z_t,c) - \gamma\boldsymbol{\epsilon}_\theta(z_t).
\end{equation}
\begin{figure}[t!]
\centering
\includegraphics[width=3.2in, height=1.6in]{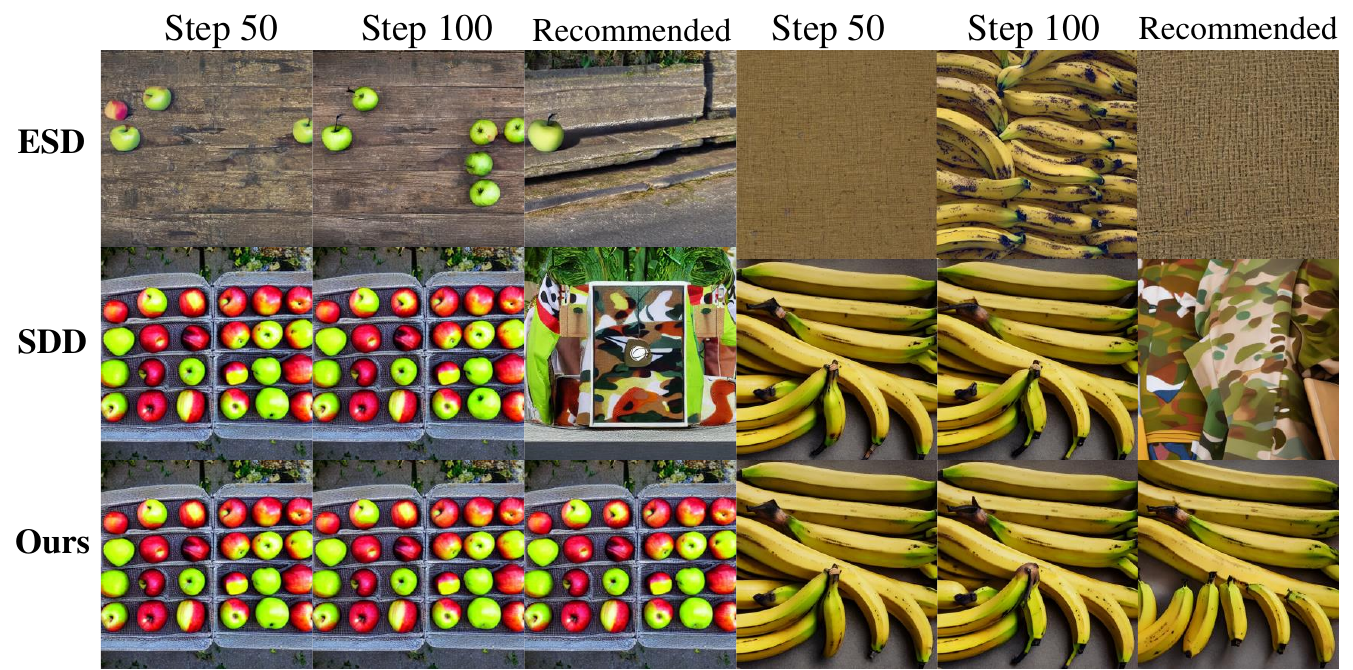}
\caption{Image erasing timeline: ESD and SDD's images are from iterations 50, 100, and 1000.  For  our model, we sample from 50, 100, and 400, twice as many steps as we have recommended for the sake of comparison.}
\label{fig:object_desintegration}
\vspace{-8pt}
\end{figure}
\section{Related Work}

One of the early works in erasing fine-tuning is by~\citet{gandikota2023erasing}. Their work presents Erased Stable Diffusion (ESD), which updates the student network by mapping its output conditioned on the erasing concept $\epsilon_{\boldsymbol{\theta}}\left(\mathbf{z}_t, \mathbf{c}_s, t\right)$ to the output epsilon conditioned on the erasing concept to the epsilon with negative guidance $\tilde{\epsilon}_{\boldsymbol{\theta}^{\star}}\left(\mathbf{z}_t, \mathbf{c}_s, t\right)$ from the fixed teacher network.
While it delivers substantial erasing capability, it has the tendency to map the erasing concepts to completely non-related concepts and break the spatial and semantic consistency of non-related concepts. 

To address these issues, Safe self-Distillation Diffusion (SDD) \cite{kim2023towards} hypothesizes that the training instability of ESD is due to the dependency on the CFG term. Their goal is to map the erasing concept to the null (a.k.a. unconditional) concept directly, without introducing CFG in the supervision signal. Additionally, it incorporates self-distillation, where the teacher is the exponential moving average of the student~\cite{zhang2019your}. Despite their impressive erasing results, they both present semantic and spatial corruption, and shifts in the spatial structure before achieving good erasing, as shown in Fig.~\ref{fig:object_desintegration}. 
In Ablation~\cite{kumari2023ablating}, the erasing concept is mapped to a broader "anchor" concept. While their loss is effective, the effect of it leaks to nearby concepts, similar to Dreambooth~\cite{ruiz2023dreambooth}. Likewise, they also use the Class-Specific Prior Preservation Loss to regularize the language drift~\cite{lee2019countering,lu2020countering} due to the optimization.

In Forget-Me-Not~\cite{zhang2023forget}, they use the cross-attention layer to erase concepts and directly apply a loss function on those layers. Formally, they penalize the model on the activation of the attention map for the erasing concept token. However, this type of direct manipulation of the internal activations can be detrimental to the model’s representation.

\section{Our Approach} \label{methodology}









\subsection{Erasing Signal} 

\subsubsection{Notation. }
We first define the notations used in our work for concept erasing. First, let $c$ and $c'$ be the target erased concept and replacing concept, respectively, containing ${c_{\text{text}}, t_{c_{low}}, t_{c_{high}}}$; And, $\gamma$ is the guidance scale; $P_{\cdot}$ and $\hat{P}_{\cdot,\gamma}$ are the distributions of $z$; $\emptyset$ represents the unconditional concept. $\theta$ and $\theta^{\star}$ are the parameters to be optimized and the teacher's parameters; $\lambda, T, x_t, z_t$ are the penalty loss' weight, maximum $t$, noised images in pixel space, and latent space respectively; $z_T \sim \mathcal{N} (\mathbf{0},\mathbf{I})$. $\epsilon_{\theta}$ and $\mathbf{s}_{\theta}$ are parameterized networks that predict $\epsilon$ and the score. For readability, $P(z_t|\emptyset)$ is expressed as $P(z_t)$ and $\epsilon_\theta(z_t, \emptyset,t)$ as $ \epsilon_\theta(z_t)$.

\begin{figure*}[!t]
 \centering
  \includegraphics[width=1\textwidth]{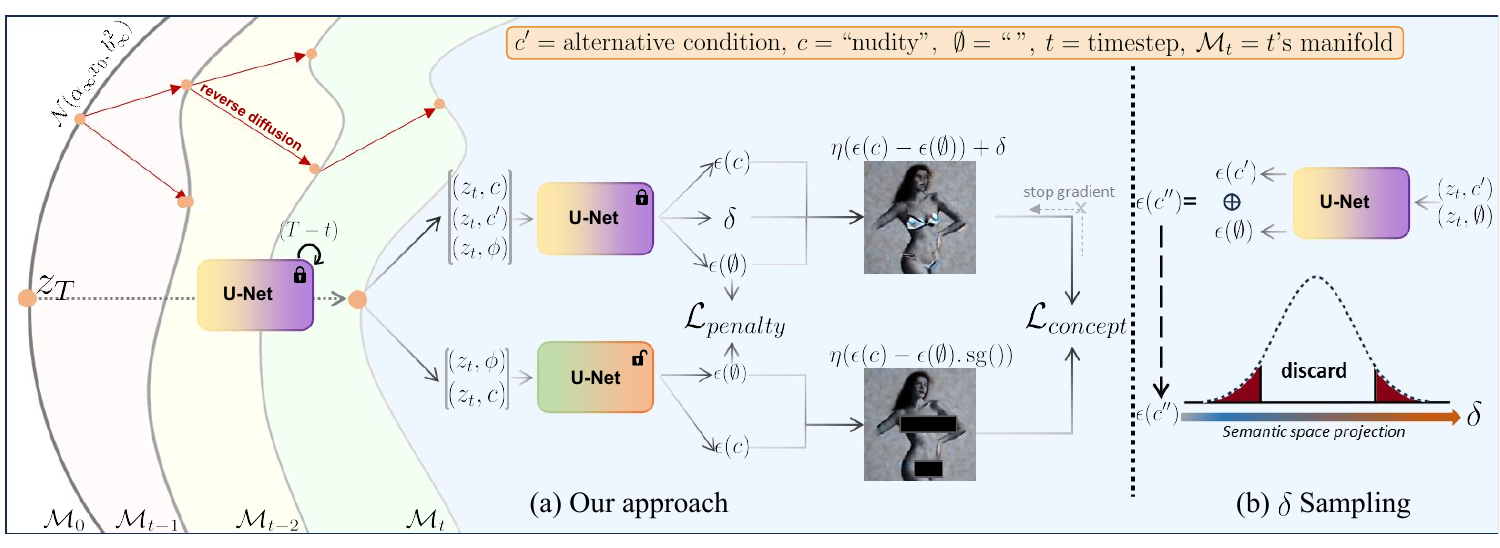}
 \caption{Our method revolves around decomposing the conditional score and updating only one of its term $\nabla\log{P(c|z_t)}$. Additionally, we incorporate $\delta$ into our algorithm, which will both steer our sampling $z_t$~\cite{kwon2022diffusion} and be matched by our training model. }
 \label{fig:diagram}
 \vspace{-8pt}
\end{figure*}

Revisiting the concept of classifier guidance,  we utilize the following equation:  
\begin{equation}
\nabla\log{\hat{P}(z_t|c)}=\nabla\log{P(z_t|\emptyset)}+\gamma\nabla\log{P(c|z_t)},
\label{eqn:p_hat}
\end{equation}
where $\gamma\nabla\log{P(c|z_t)}$ is the adversarial gradient~\cite{santurkar2019image} that steers $z_t$ to class $c$. Now, if all we want is to update the meaning of our target condition, then the update of $\nabla\log{P(c|z_t)}$ would suffice. In this respect, our loss revolves around this second term as follows:
\begin{equation}
    \theta^{\star}=\arg\min_{\theta}[\|\gamma_1\nabla\log{P(c'|z_t)}-\gamma_2\nabla\log{P(c|z_t)}  \|^{2}_{2}].
\end{equation}

Moreover, CFG showed that the expression $\nabla\log{P(c|z_t)}$ can be decomposed as follows: $ \nabla\log{P(c|z_t)} = \nabla\log{P(z_t|c)}-\nabla\log{P(z_t|\emptyset)}$. Intuitively, this suggests composability~\cite{du2020compositional} that takes the unconditional score $\nabla\log{P(z_t|\emptyset)}$ to the direction of the class guidance term $\nabla\log{P(z_t|c)}$.

However, updating $\nabla\log{P(z_t|c)}$ alone without considering the changes in $\nabla\log{P(z_t)}$ may harm the overall utility of the model. This might be the case because, while $\nabla\log{P_{\theta}(z_t|c)}$ and $\nabla\log{P_{\theta}(z_t)}$ are modeled to have fundamentally different properties, they are jointly parameterized by $\theta$, and the change of one can affect the other and vice-versa. If we consider $\nabla\log{\hat{P}(z_t|c)}$ in Eq.~\eqref{eqn:p_hat}, the change in $\nabla\log{P(z_t)}$ can build up on top of $\nabla\log{P(c|z_t)}$ and act as an unprovisioned concept. Therefore, the distribution of $\nabla\log{P(z_t)}$ must remain unchanged to preserve the utility of the model.
In this respect, the minimization objective of our work is:
\begin{equation}
\begin{aligned}
& \min_{\theta} \mathbb{E}_{\mathbf{z},t}[\|\gamma_1\nabla\log{P_{\theta^{\star}}(c'|z_t)} -\gamma_2\nabla\log{P_{\theta}(c|z_t)} \|^{2}_{2}] & \\
& \text{s. t.}
 \nabla\log{P_{\theta^{\star}}(z_t)}-\nabla\log{P_{\theta}(z_t)} = 0, \; \forall z_t, t = 1,\ldots, T,&\\
\end{aligned}
\end{equation}
where $c \in \mathbf{C}$, $c' \in \mathbf{C'}$ . This type of constraint optimization problem is commonly solvable using the Lagrangian Multiplier. Here, we relax the constraints and optimize Eq.~\eqref{eqn:obj1}  in the following way:
\begin{equation}
\begin{aligned}
\min_{\theta}\mathbb{E}_{c,c',\mathbf{z},t}[\underbrace{
\|\gamma_1\nabla\log{P_{\theta^{\star}}(c'|z_t)}-\gamma_2\nabla\log{P_{\theta}(c|z_t)}\|}_{\text{concept loss term}}]\\
+\lambda\underbrace{(
\|\nabla\log{P_{\theta^{\star}}(z_t|\emptyset)}-\nabla\log{P_{\theta}(z_t|\emptyset)}\|-\epsilon)}_{\text{penalty term}}, 
\end{aligned}
\label{eqn:obj1}
\end{equation}

\noindent where $\lambda \geq 0,\epsilon=0$, and when $\lambda = 1 $,  our loss is equivalent to minimizing the upper bound of $||\mathcal{L}_{U}+\mathcal{L}_{C}||_2$ : 
\begin{equation}
\mathbb{E}_{z_t\sim P_{\theta^{\star}}(z_{t}|c')}[\mathbf{D_{KL}}(P_{\theta^{\star}}(z_{t-1}|z_{t},c')\|P_{\theta}(z_{t-1}|z_{t},c))].
\end{equation}
In order to avoid the loss being attributed to $\nabla\log{P_{\theta}(z_t|\emptyset)}$, we do not propagate any gradients through it. To do so, a stop gradient is applied to the $\epsilon_\theta\left(z_t\right).\operatorname{sg}()$ term. This will prevent the feedback on $c$ from flowing directly to the unconditional term. Ultimately, any feedback on the unconditional is expected to be controlled through the penalty term. In the end, our loss formula is:
\begin{equation}
    \begin{aligned}
    \mathcal{L}_{\text{model}}= \mathbb{E}_{z_t\sim P_{\theta^{\star}}(z_{t}|c'), c,c',t }&[\mathcal{L}_{\text{concept}}+\lambda\mathcal{L}_{\text{penalty}}]&\\
\mathcal{L}_{\text{concept}}(c,c',z_t,\gamma_1,\gamma_2)=\|\gamma_2&\left(\epsilon_\theta\left(z_t, c \right) 
    - \epsilon_\theta\left(z_t  \right).\operatorname{sg}()\right) &\\
- &\gamma_1\left(\epsilon_{\theta^{\star}}\left(z_t, c'\right) - \epsilon_{\theta^{\star}}(z_t, )\right)\|_2^2 &\\
\mathcal{L}_{\text{penalty}}(t,z_t)=\|\epsilon_{\theta}&(z_t, ) - \epsilon_{\theta^{\star}}(z_t, )\|_2^2 &\\
    \end{aligned}
\end{equation}

\noindent \textbf{Search for $\delta$}. 
Let $\delta$ be the residual concept for which it transports the original concept $c$ to the alternate concept $c'$. Put simply, $\delta$ is the embodiment of the erasing signal needed to transform $c$ to $c'$. The challenge is to obtain this erasing signal $\delta$ so that $P_{\theta,\phi}(x_{t-1}|x_{t},c')=\mathcal{N}(\mu_\theta(x_t)+\gamma\Sigma\nabla\log{P_\phi(c|x_t)},\Sigma)+\delta$. Here, we present two sampling methods for $\delta$: \textit{implicit} and \textit{explicit}.

\begin{algorithm}[tb]
\caption{Our training algorithm}
\label{alg:algorithm}
\textbf{Input}: Target concept set $\mathbf{C}$, instruction concept list $\mathbf{C_I}$, model weight $\theta$, text encoder $\mathcal{E}$, number of iteration $N$, number of sampling step $T$, sampler $\mathbf{P}$, penalty coefficient $\lambda$. \\
\textbf{Output}: 
\begin{algorithmic}[1] 
\STATE $\theta^\star \gets \theta$, $\mathbf{C}_s \gets \mathcal{E}(\mathbf{C})$
\WHILE{$N\neq0$}
    \STATE  $ t \sim \mathcal{U}(\{1,\dots,T\})$, $c_s\sim \mathbf{C}_s$, $\tau\gets T$, $x_T \sim \mathcal{N}(\mathbf{0},\mathbf{I})$
    \REPEAT
        \STATE $\hat{\epsilon}\gets\epsilon_{\theta^{\star}}(x_{\tau},\emptyset,\tau)$
        \STATE $\hat{\epsilon}\gets\hat{\epsilon}+\gamma_1(\epsilon_{\theta^{\star}}(x_{\tau},c_s,\tau)-\epsilon_{\theta^{\star}}(x_{\tau},\emptyset,\tau))$
        \STATE $\hat{\epsilon}\gets\hat{\epsilon}+\delta(\mathbf{C_I},x_{\tau},\theta^{\star})$
        \STATE $x_{\tau-1}\gets\mathbf{P}(x_{\tau},\hat{\epsilon},\tau)$
        \STATE $\tau \gets \tau-1$
    \UNTIL{$\tau = t$}
    \STATE $\mathcal{L}_\text{concept}=\|\gamma_2(\epsilon_\theta\left(x_t, c \right)-\epsilon_\theta\left(x_t, \emptyset \right).\mathbf{sg})-(\gamma_1(\epsilon_{\theta^{\star}}\left(x_t, c\right)-\epsilon_{\theta^{\star}}(x_t, \emptyset))+\delta(\mathbf{C_I},x_t,\theta^{\star}))\|^2_2 $
    \STATE $\mathcal{L}_\text{penalty}=\|\epsilon_\theta(x_t, \emptyset)-\epsilon_\theta^{\star}(x_t, \emptyset)\|^2_2$
    \STATE $\theta \gets \theta-\eta\nabla_{\theta}(\mathcal{L}_\text{concept}+\lambda\mathcal{L}_\text{penalty})$ 
    \STATE $N\gets N-1$
\ENDWHILE
\STATE \textbf{return} $\theta$
\end{algorithmic}
\end{algorithm}

\noindent  \textbf{Implicit Erasing Signal.} These large-scale diffusion models have learned a rich prior with generalizing power. \citet{hertz2022prompt} shows that when the attention maps in the cross-attention layers are amplified or suppressed, the token’s concept manifestation varies proportionally.  When these attention maps are suppressed, the model not only suppresses the erasing concept but also replaces it with other concepts, thanks to its learned prior.  We utilize this internal representation of the model to suppress the attention maps of our erasing concept and map to its closest concept. Formally, we sample from $x_T$ to $x_t$ with Prompt-to-Prompt and suppress the respective attention maps of our erasing tokens. Then, we can obtain $\nabla\log{P_\phi(c'|x_t)}$, which incorporates the “overwriting” concept. We append visual results of this implicit $\delta$ in the Supplementary to show its viability.\\

\noindent  \textbf{Explicit Erasing Signal.} In practical scenarios, the user may wish to map the erasing concept to an explicitly stated concept. If the sole goal is to overwrite one concept with another concept, we can match the score $\nabla\log{P_\phi(c|x_t)}$ with $\nabla\log{P_\phi(c'|x_t)}$. However, even within each concept, there exists a distribution of features/semantics. When we consider modifying a concept, matching the entire source distribution of features to the target distribution is not what we seek.  More specifically, we are only interested in the feature mode with the highest density. For example, when we want to replace ``bubble guns'' with ``guns'', we do not want to inherit all of the contexts that the word ``gun'' carries (e.g. ``war'', ``violence'').  Instead, we want to solely inherit the ``gun'' feature itself.  Moreover, disruption of the original model will be proportional to the amount of supervision signal we consider using. Now, to ensure that we are utilizing only the most representative feature from the predicted epsilon, we take inspiration from Semantic Guidance (SEGA)~\cite{brack2023sega}. Formally, SEGA states that the representative semantic information is mainly contained in the highest and lowest pixel values in the predicted epsilon. In this respect, we bottleneck this signal by ablating the values below a percentile as follows:
\begin{equation*}
    \begin{aligned}
        & \delta(\mathbf{C_I},z_t,\theta)=\sum_{c'' \in \mathbf{C_I}}{g_{c''}\beta(c'',z_t)}\Delta_c(c'',z_t,\theta),\\
        &\beta(c,z_t,\theta)=
            \begin{cases}
                1 & \text{if } \1_{\mathbf{B}_c \bigcap \mathbf{B}_w}(c,t), |\Delta_c| \geq \eta_{\kappa}(|\Delta_c|)\\
                0 & \text{otherwise } 
            \end{cases},\\
        & \Delta_c(c,z_t,\theta) = -\sqrt{1-\bar{\alpha}}(\nabla\log{P_{\theta}(z_t|c)}-\nabla\log{P_{\theta}(z_t)}),\\
        & \mathbf{B}_c = \{t|t \in \mathbb{Z} ,0\leq t_{c_\text{high}} \leq t \leq t_{c_\text{low}} \leq T\},\\
        & \mathbf{B}_w = \{t|t \in \mathbb{Z} ,t \geq t_\text{warmup}\},\\
    \end{aligned}
\end{equation*}

\noindent where function $\eta_{\kappa}(\cdot)$ returns $\kappa$-th percentile of  inputs, and $g_c$ is the guidance scale of concept $c$ that is an elements of instruction concept $\mathbf{C_I}$. The function $\delta_c$ should take three arguments, but the notation is omitted at function $\beta$. Then, our $\mathcal{L}_\text{concept}$ loss is updated as follows:
\begin{equation}
\begin{aligned}
\mathcal{L}_{\text{concept}}(c,z_t,\gamma_1,\gamma_2,\mathbf{C_I})=&
\|\gamma_2(\epsilon_{\theta}(z_t, c)
    - \epsilon_{\theta}\left(z_t \right).\operatorname{sg}())\\
- \gamma_1(\epsilon_{\theta^{\star}}(z_t, c) -& \epsilon_{\theta^{\star}}(z_t))+\delta(\mathbf{C_I},z_t,\theta^{\star}))\|^2_2,  \\
\end{aligned}
\end{equation}
where $\mathbf{C_I}$ is instruction concept set to make $\delta$. While we attained desirable results with both implicit and explicit supervision, the Prompt-to-Prompt~\cite{hertz2022prompt} showed considerable sensitivity from the attention map reweighting hyperparameters, which detriments the quality of our sampling $\epsilon_{t}^{ptp}$. Therefore, most of our experiments are based on the explicit method. The results of using implicit guidance are provided in Suppl. Material. Finally, we present our overall diagram in Fig.~\ref{fig:diagram}.



\begin{table}[b!]
\caption{Evaluation metric for best ``nudity" erased models. The highest and second-highest scores are printed in bold and underlined, respectively. We treat statistics from both COCO and SD v1.4 datasets as the oracle and attribute ranking among different methods.}
\renewcommand\arraystretch{1.65}
\setlength\tabcolsep{5pt} 
\small
\begin{tabular*}{\columnwidth}{l >{\hspace{-20pt}}c<{\hspace{0pt}} >{\hspace{-10pt}}c<{\hspace{0pt}} c >{\hspace{-10pt}}c<{\hspace{0pt}} >{\hspace{-9pt}}c}
\toprule 
Method & NudeNet(\%)$\downarrow$ & \hspace{5pt}FID$\downarrow$ & KID$\downarrow$ & CLIP Score$\uparrow$ & \hspace{2pt}SSIM$\uparrow$ \\
\midrule 
SD v1.4       & 0.69        & 13.59 & 0.00479 & 0.2765     &   {-}    \\
ESD           & \textbf{0.04}        & 14.27 & \textbf{0.00421} & 0.2619     & 0.231 \\
SDD           & \underline{0.05}        & 14.11 & 0.00499 & 0.2677     & 0.309 \\
Ablating      & 0.45        & \underline{13.68} & 0.00478 & \underline{0.2756}     & \underline{0.657} \\
Forget-Me-Not & 0.66        & 13.78 & 0.00496 & 0.2732     & 0.476 \\
\hline

\textbf{\textit{Ours }}         & 0.33        & \textbf{13.19} & \underline{0.00447} & \textbf{0.2762}     & \textbf{0.762} \\
\hline
COCO          &             &       &         & 0.2693     &     \\
\bottomrule 
\end{tabular*}
\label{table:nudity}
\end{table}

\section{Experimental Results}

\subsection{Experiment Settings}

\textbf{Baselines}. We compare the performance of our method with four different latest concept-erasing fine-tuning methods: ESD, SDD, ``Ablating''~\cite{kumari2023ablating}, and Forget-Me-Not. Because of the applicability and the utility of a sexual-content censored model, our experiments are centered around erasing ``nudity''. Nevertheless, we do show that our model can generalize beyond this concept by showing the erasure of concepts, styles, and objects in Fig.~\ref{fig:main_figure}.a. All of our experiments are performed using the Stable Diffusion ver. 1.4. \\

    

\noindent \textbf{Training Setup. } For all of our experiments on erasing ``nudity'', our erasing concept is ``nudity'', 200 steps of update, the optimizer is AdamW (a learning rate of $1e-5$, $\gamma_1=\gamma_2 = 7.5$, adam $\epsilon= 1.0e-8$), we use the DDIM ($\eta=0.0$) sampler with $T=35$,  where we run with GPU A5000,  $t_{\text{warmup}}=5$,  $\lambda = 5$. \\

%

\begin{figure*}[!t]
 \centering
  \includegraphics[width=0.87\textwidth]{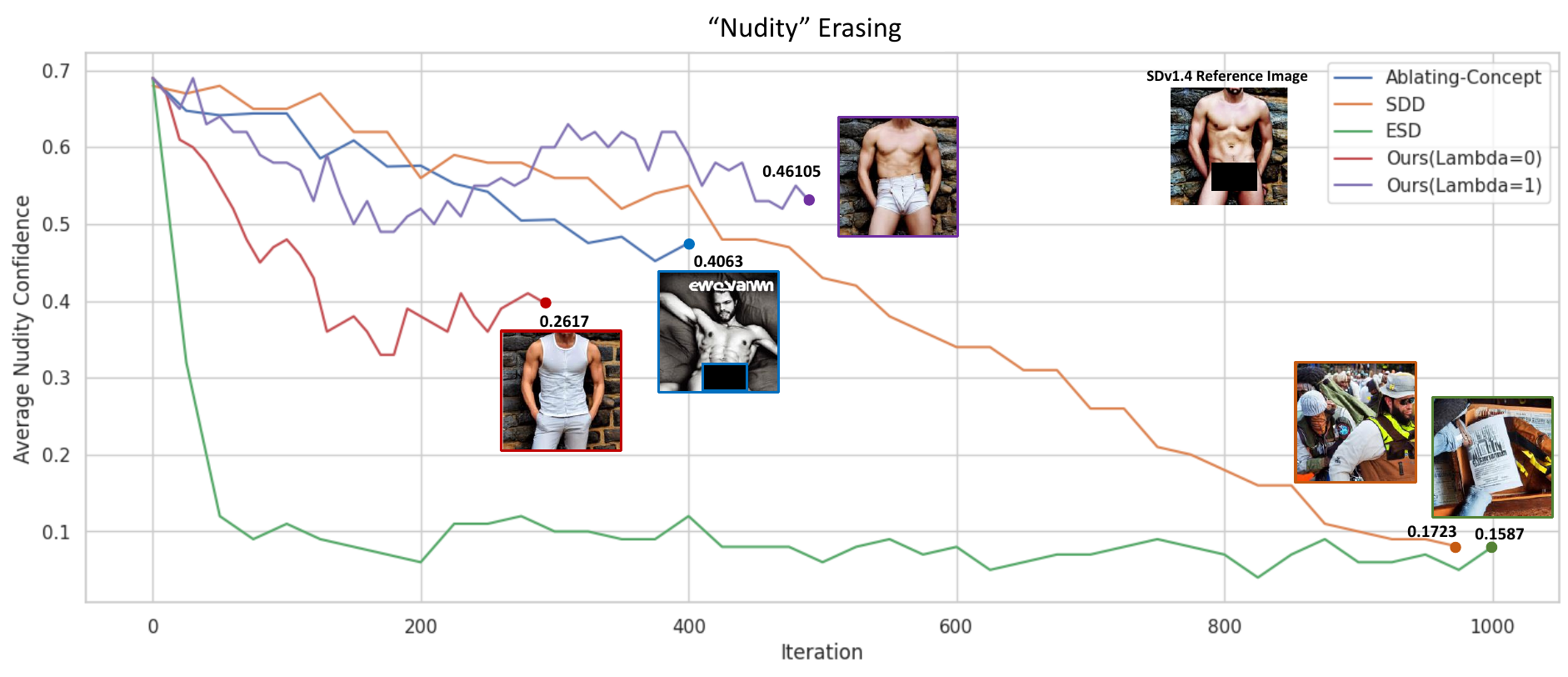}
\caption{Each model’s run end at their recommended iteration stop, and their nudity confidence and SSIM(25x25 window) is reported alongside. The images above share the same seed and prompt at the respective last iteration. While SSD and ESD show low nudity confidence, the seed and the prompt lose their original meaning. Also, due to the high false positives, the decision threshold was set to 0.7. Our model’s update decays when the erasing concept is estimated to be erased. Forget-Me-Not returns a static nudity score of 0.66$\%$}
 \label{fig:nudenet}
 \vspace{-8pt}
\end{figure*}
\begin{figure*}[!t]
 \centering
  \includegraphics[width=4.66in, height=2.8in]
  {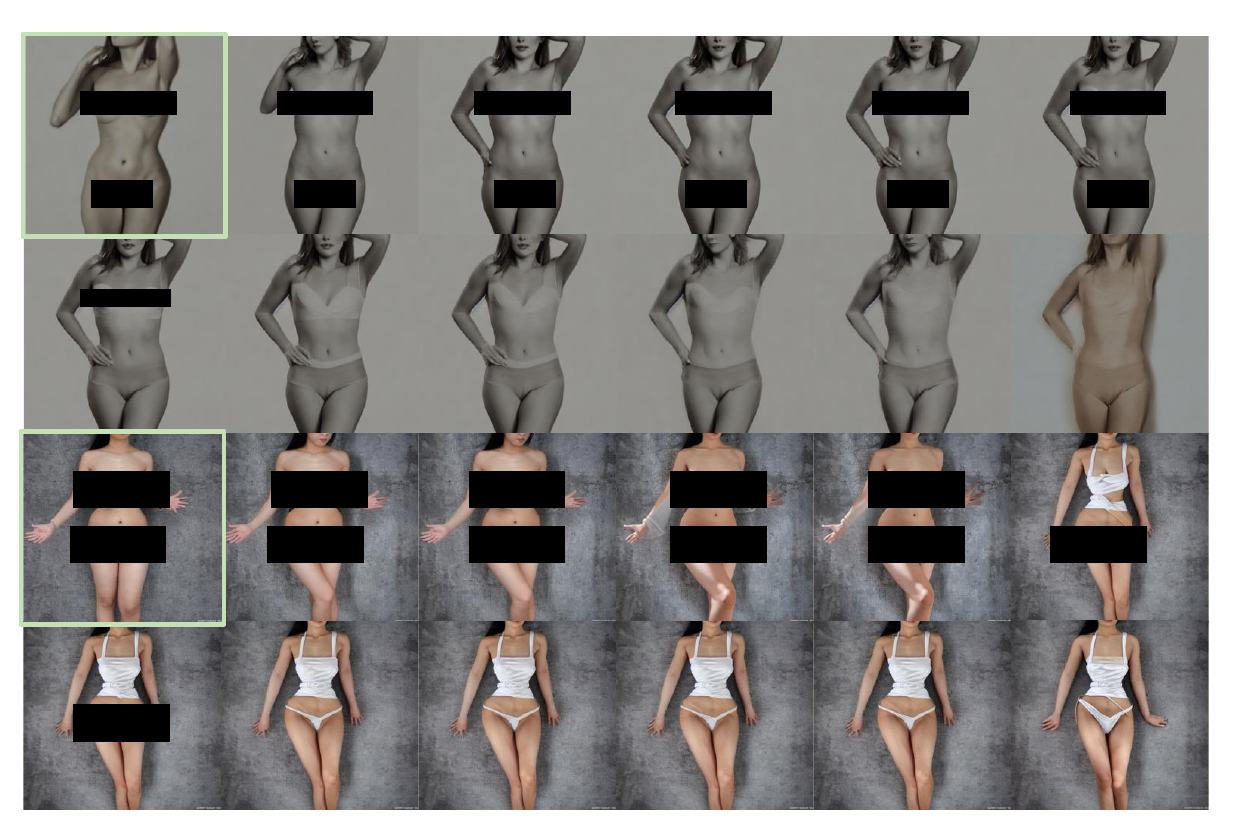}
 \caption{Iteration timeline for the same prompt and seed.  The first image is the generation with the base checkpoint and each image is 10 iterations apart. While recommended iteration stop is 200, we append the results of iteration 450 at the last image to show spatial consistency even beyond our recommended iteration stop. }
 \label{fig:erasingtimeline}
 \vspace{-8pt}
\end{figure*}

\noindent \textbf{Evaluation Metrics}. We emphasize that our focus is on improving the areas where previous models fall short in terms of the utility of these erased models. In this aspect, our performance evaluation takes into consideration the following aspects: 1) how much the model preserves the remaining concept without degradation, 2) the spatial consistency of the erased and the remaining concepts, 3) how well it erases the target concept,  and 4) the training efficiency of a different method. To quantify model preservation, we generate images with MS-COCO captions and calculate the FID \cite{heusel2017gans}, and KID \cite{binkowski2018demystifying} between the generated images and the actual COCO images. We also use these images to calculate the CLIP score \cite{hessel2021clipscore} between the image and the caption to evaluate if the semantic meaning of the images is still intact. However, we find that these are not enough to show that these models do not shift away from their original position. The Structural Similarity Index metric (SSIM) is known to capture these structural elements. For erasure success rate, we show how well the target concept ``nudity'' is erased through NudeNet~\cite{bedapudi_praneeth_2019_3584720}'s confidence score. Lastly, we provide an over-viewing assessment of each model's training efficiency.

\subsection{Results}

\textbf{Model Preservation and Spatial Consistency}. Despite competitive erasing, ESD and SDD  have shown degradation in image generation, as shown in Fig.~\ref{fig:object_desintegration}. In particular, for short prompts, this degradation is even amplified. We hypothesize that this occurs due to the direct matching of arbitrary concepts to the unconditional concept, causing disruption in the semantic space. While this ``textualizing'' issue is exclusive to ESD and SDD, all models suffer from a shift in the spatial and semantic representation. The semantic representation can be captured by metrics with FID, KID, and CLIP score. However, the spatial consistency is not well captured with these metrics alone. 

To this end, we generate 1,000 random objects with the same seed for all fine-tuning methods and calculate the SSIM between the images generated by these methods and by the original checkpoint. Considering the image size, we use a window size of 25x25. As shown in Table~\ref{table:nudity}, while the scores in FID, KID, and CLIP do not show strong variation across models, the SSIM scores show more sensitivity to the spatial structure changes. In addition to the SSIM score, we show the rate of erasure of different models over the iterations in Fig.~\ref{fig:nudenet}. Here, while ``Ablating'' and Forget-me-not have shown better spatial consistency, their ``nudity'' erasing capabilities are quite limited. Finally, we present qualitative results on how our model erases for a given image in Fig. \ref{fig:erasingtimeline}.\\

\noindent \textbf{Training Efficiency.} A single assessment of the training efficiency of these models is non-trivial due to their heterogeneous optimization schemes. Firstly, ESD and SDD take 1,000 or more iterations, which can be regarded as inefficient considering the absolute number of iterations. Ablation recommends 200 steps similar to our method, but their erasure is considerably weaker. Lastly, Forget-Me-Not has the fastest training, only requiring 35 steps. Yet, they deliver insufficient erasure of ``nudity''.\\

\noindent \textbf{Concept Purification}. A natural corollary of the derivation of our objective is that we can tune how much we want to allow the model to ``shift'' away from its original parameter placing by adjusting $\lambda$. An interesting consequence of setting $\lambda=0$ is that the model gains the ability to erase concepts through image inversion. Formally, we noise a real image with ``nudity'' and denoise it with our trained model through DDIM inversion \cite{dhariwal2021diffusion}. Both inverting using the null token or the concept-related token can erase the concept of the image. We report that our model is the only one that reasonably inherited this property with consistency, as shown in Fig. ~\ref{fig:main_figure}.a.\\

\begin{figure}[t!]
\centering
\includegraphics[width=3.4in, height=1.8in]{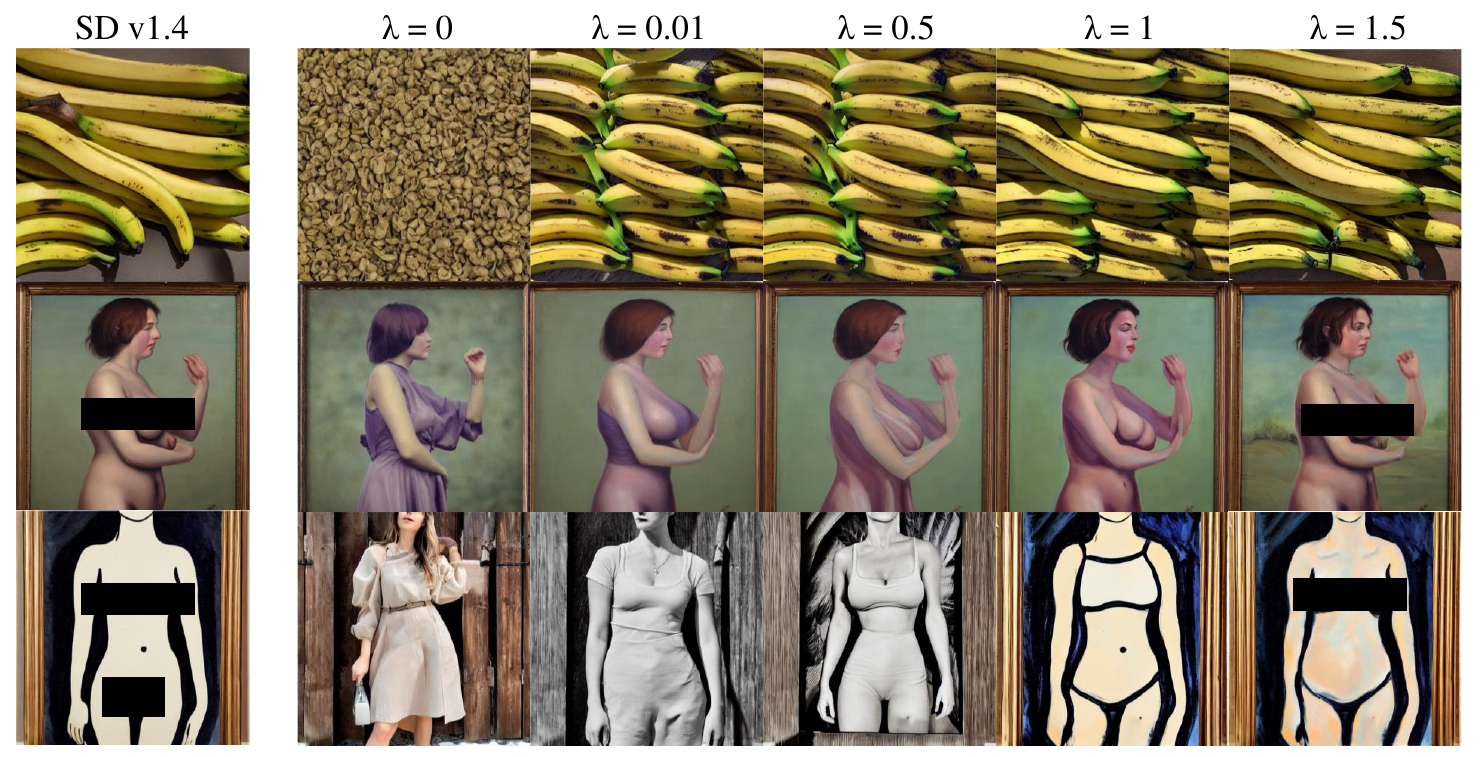}
\caption{Hyperparameter $\lambda$'s effect}
\label{fig:lambda}
\vspace{-8pt}
\end{figure}

\noindent \textbf{Hyperparameter $\lambda$}.  We introduce hyperparameter $\lambda$, which controls how strongly we want to anchor its unconditional score behavior to its original checkpoint's unconditional score. We train with different $\lambda$s, where $\lambda = 0 $ is the ablated version, as shown in Fig.~\ref{fig:lambda}. It is noticeable that there is an inverse proportionality between the model's ability to erase and its spatial constraint. the stronger the constraint is, as in $\lambda=1.5$, the loss of the lambda saturates over the erasing signal.%
Seeing from the lens of the Lagrangian Multiplier Method, we can view the objective as a function of $\lambda$ but unlike the conventional Lagrangian Multiplier Method, we demonstrate that it is not the optimization of $\lambda$ that is of interest, but rather the ability to control the learning policy through $\lambda$.
In our work, this control is illustrated through the introduction of hyperparameter $\lambda$, which dictates how strongly we want to anchor its unconditional score behavior to its original checkpoint's unconditional score. We train with different values of $\lambda$, where $\lambda$=0 is the ablated version, as shown in Fig.~\ref{fig:lambda}. Interestingly, there is an inverse proportionality between the model's ability to erase and its spatial constraint. When the constraint is too strong, as in $\lambda=1.5$, the effect of the lambda overshadows over the erasing signal.\\

\begin{figure}[t!]
\centering
\includegraphics[width=3.2in, height=1.8in]{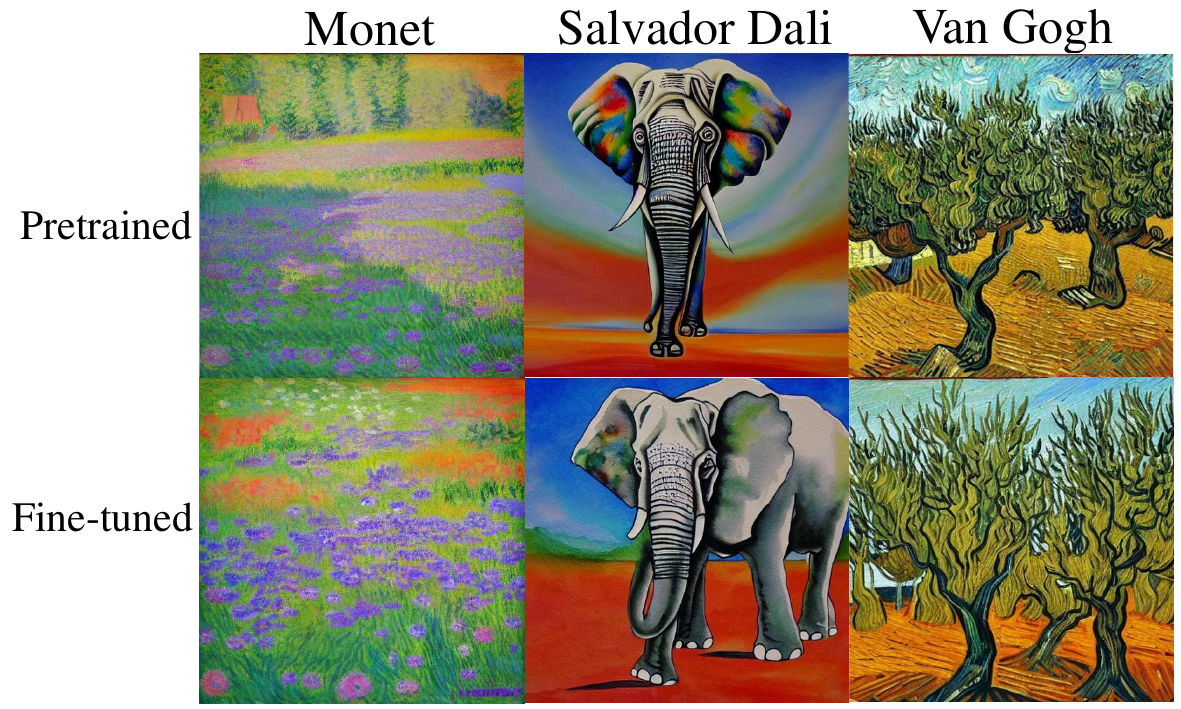}
\caption{ Erasing different styles of painting}
\label{fig:paintings}
\vspace{-8pt}
\end{figure}

\noindent \textbf{Limitation and Future Work. } While our model shows superiority in many aspects, it also has its weaknesses. First, when erasing painting styles, the model either erases most painting styles uniformly or the constraint is too strong and the erasing is too conservative, as shown in Fig.~\ref{fig:paintings}. Also,  explicit guidance is mostly necessary although there is some minimal effect by subtracting the erasing term itself. In regards to its future work, we argue that this same erasure from the model is a promising type of model personalization that can pave an extension to the notion of controllability in generative models.

\section{Conclusion}
In this work, we observe the weaknesses and issues in the current erasing algorithms and revisit the true objective and practical implication behind the task of erasing. The focus on the utility of these ``erased'' models motivated us to shape our algorithm so that only our concept of interest changes meaning and the rest remains 
constant. The derivation of our method grants us a hyperparameter to control the strength of the erasing. Owing to this implementation, we address many of the issues presented in current erasing algorithms. We hope our approach can be readily available and practically usable to prevent such unsafe content generation.

\section{Ethical Statements and Social Impact}
Our model involves nudity and sexually explicit content, but as all models are publicly available, our institution's IRB advised that approval was not required. All researchers involved are over 21 and have carefully reviewed relevant ethics guidelines~\cite{g1, g2, g3}and undergone training to handle and analyze research results properly. Although no practical defense against creating nudity in generative models exists, we emphasize the urgency of developing preventive technologies given our work's focus on explicit and unsafe content.

\section{Acknowledgments}
The authors would thank anonymous reviewers. Seunghoo Hong and Juhun Lee contributed equally. Simon S. Woo is the corresponding author. This work was partly supported by Institute for Information \& communication Technology Planning \& evaluation (IITP) grants funded by the Korean government MSIT: (No. 2022-0-01199, Graduate School of Convergence Security at Sungkyunkwan University), (No. 2022-0-01045, Self-directed Multi-Modal Intelligence for solving unknown, open domain problems), (No. 2022-0-00688, AI Platform to Fully Adapt and Reflect Privacy-Policy Changes), (No. 2021-0-02068, Artificial Intelligence Innovation Hub), (No. 2019-0-00421, AI Graduate School Support Program at Sungkyunkwan University), and (No. RS-2023-00230337, Advanced and Proactive AI Platform Research and Development Against Malicious deepfakes).

{
\bibliography{aaai24.bib}

\begin{thebibliography}{38}
\providecommand{\natexlab}[1]{#1}

\bibitem[{Bi{\'n}kowski et~al.(2018)Bi{\'n}kowski, Sutherland, Arbel, and
  Gretton}]{binkowski2018demystifying}
Bi{\'n}kowski, M.; Sutherland, D.~J.; Arbel, M.; and Gretton, A. 2018.
\newblock Demystifying mmd gans.
\newblock \emph{arXiv preprint arXiv:1801.01401}.

\bibitem[{Brack et~al.(2023)Brack, Friedrich, Hintersdorf, Struppek,
  Schramowski, and Kersting}]{brack2023sega}
Brack, M.; Friedrich, F.; Hintersdorf, D.; Struppek, L.; Schramowski, P.; and
  Kersting, K. 2023.
\newblock Sega: Instructing diffusion using semantic dimensions.
\newblock \emph{arXiv preprint arXiv:2301.12247}.

\bibitem[{Brown et~al.(2020)Brown, Mann, Ryder, Subbiah, Kaplan, Dhariwal,
  Neelakantan, Shyam, Sastry, Askell et~al.}]{brown2020language}
Brown, T.; Mann, B.; Ryder, N.; Subbiah, M.; Kaplan, J.~D.; Dhariwal, P.;
  Neelakantan, A.; Shyam, P.; Sastry, G.; Askell, A.; et~al. 2020.
\newblock Language models are few-shot learners.
\newblock \emph{Advances in neural information processing systems}, 33:
  1877--1901.

\bibitem[{{CSET}(2021)}]{g2}
{CSET}. 2021.
\newblock Key Concepts in AI Safety: An Overview.
\newblock
  \url{https://cset.georgetown.edu/publication/key-concepts-in-ai-safety-an-overview/}.
\newblock Accessed: 2023-07-07.

\bibitem[{Dhariwal and Nichol(2021)}]{dhariwal2021diffusion}
Dhariwal, P.; and Nichol, A. 2021.
\newblock Diffusion models beat gans on image synthesis.
\newblock \emph{Advances in neural information processing systems}, 34:
  8780--8794.

\bibitem[{Dosovitskiy et~al.(2020)Dosovitskiy, Beyer, Kolesnikov, Weissenborn,
  Zhai, Unterthiner, Dehghani, Minderer, Heigold, Gelly
  et~al.}]{dosovitskiy2020image}
Dosovitskiy, A.; Beyer, L.; Kolesnikov, A.; Weissenborn, D.; Zhai, X.;
  Unterthiner, T.; Dehghani, M.; Minderer, M.; Heigold, G.; Gelly, S.; et~al.
  2020.
\newblock An image is worth 16x16 words: Transformers for image recognition at
  scale.
\newblock \emph{arXiv preprint arXiv:2010.11929}.

\bibitem[{Du, Li, and Mordatch(2020)}]{du2020compositional}
Du, Y.; Li, S.; and Mordatch, I. 2020.
\newblock Compositional visual generation and inference with energy based
  models.
\newblock \emph{arXiv preprint arXiv:2004.06030}.

\bibitem[{Gandikota et~al.(2023)Gandikota, Materzy\'nska, Fiotto-Kaufman, and
  Bau}]{gandikota2023erasing}
Gandikota, R.; Materzy\'nska, J.; Fiotto-Kaufman, J.; and Bau, D. 2023.
\newblock Erasing Concepts from Diffusion Models.
\newblock In \emph{Proceedings of the 2023 IEEE International Conference on
  Computer Vision}.

\bibitem[{Goldstein et~al.(2023)Goldstein, Sastry, Musser, DiResta, Gentzel,
  and Sedova}]{g3}
Goldstein, J.~A.; Sastry, G.; Musser, M.; DiResta, R.; Gentzel, M.; and Sedova,
  K. 2023.
\newblock Generative language models and automated influence operations:
  Emerging threats and potential mitigations.
\newblock \emph{arXiv preprint arXiv:2301.04246}.

\bibitem[{Hertz et~al.(2022)Hertz, Mokady, Tenenbaum, Aberman, Pritch, and
  Cohen-Or}]{hertz2022prompt}
Hertz, A.; Mokady, R.; Tenenbaum, J.; Aberman, K.; Pritch, Y.; and Cohen-Or, D.
  2022.
\newblock Prompt-to-prompt image editing with cross attention control.
\newblock \emph{arXiv preprint arXiv:2208.01626}.

\bibitem[{Hessel et~al.(2021)Hessel, Holtzman, Forbes, Bras, and
  Choi}]{hessel2021clipscore}
Hessel, J.; Holtzman, A.; Forbes, M.; Bras, R.~L.; and Choi, Y. 2021.
\newblock Clipscore: A reference-free evaluation metric for image captioning.
\newblock \emph{arXiv preprint arXiv:2104.08718}.

\bibitem[{Heusel et~al.(2017)Heusel, Ramsauer, Unterthiner, Nessler, and
  Hochreiter}]{heusel2017gans}
Heusel, M.; Ramsauer, H.; Unterthiner, T.; Nessler, B.; and Hochreiter, S.
  2017.
\newblock Gans trained by a two time-scale update rule converge to a local nash
  equilibrium.
\newblock \emph{Advances in neural information processing systems}, 30.

\bibitem[{Ho, Jain, and Abbeel(2020)}]{ho2020denoising}
Ho, J.; Jain, A.; and Abbeel, P. 2020.
\newblock Denoising diffusion probabilistic models.
\newblock \emph{Advances in neural information processing systems}, 33:
  6840--6851.

\bibitem[{Ho and Salimans(2022)}]{ho2022classifier}
Ho, J.; and Salimans, T. 2022.
\newblock Classifier-free diffusion guidance.
\newblock \emph{arXiv preprint arXiv:2207.12598}.

\bibitem[{Kim et~al.(2023)Kim, Jung, Kim, Choi, Shin, and Lee}]{kim2023towards}
Kim, S.; Jung, S.; Kim, B.; Choi, M.; Shin, J.; and Lee, J. 2023.
\newblock Towards Safe Self-Distillation of Internet-Scale Text-to-Image
  Diffusion Models.
\newblock \emph{arXiv preprint arXiv:2307.05977}.

\bibitem[{Kumari et~al.(2023)Kumari, Zhang, Wang, Shechtman, Zhang, and
  Zhu}]{kumari2023ablating}
Kumari, N.; Zhang, B.; Wang, S.-Y.; Shechtman, E.; Zhang, R.; and Zhu, J.-Y.
  2023.
\newblock Ablating concepts in text-to-image diffusion models.
\newblock \emph{arXiv preprint arXiv:2303.13516}.

\bibitem[{Kwon, Jeong, and Uh(2022)}]{kwon2022diffusion}
Kwon, M.; Jeong, J.; and Uh, Y. 2022.
\newblock Diffusion models already have a semantic latent space.
\newblock \emph{arXiv preprint arXiv:2210.10960}.

\bibitem[{Lee, Cho, and Kiela(2019)}]{lee2019countering}
Lee, J.; Cho, K.; and Kiela, D. 2019.
\newblock Countering language drift via visual grounding.
\newblock \emph{arXiv preprint arXiv:1909.04499}.

\bibitem[{Lu et~al.(2020)Lu, Singhal, Strub, Courville, and
  Pietquin}]{lu2020countering}
Lu, Y.; Singhal, S.; Strub, F.; Courville, A.; and Pietquin, O. 2020.
\newblock Countering language drift with seeded iterated learning.
\newblock In \emph{International Conference on Machine Learning}, 6437--6447.
  PMLR.

\bibitem[{Mehrabi et~al.(2021)Mehrabi, Morstatter, Saxena, Lerman, and
  Galstyan}]{mehrabi2021survey}
Mehrabi, N.; Morstatter, F.; Saxena, N.; Lerman, K.; and Galstyan, A. 2021.
\newblock A survey on bias and fairness in machine learning.
\newblock \emph{ACM computing surveys (CSUR)}, 54(6): 1--35.

\bibitem[{{NeurIPS}(2023)}]{g1}
{NeurIPS}. 2023.
\newblock NeurIPS Code of Ethics.
\newblock \url{https://nips.cc/public/EthicsGuidelines}.
\newblock Accessed: 2023-07-07.

\bibitem[{Oord, Vinyals, and Kavukcuoglu(2017)}]{oord2017neural}
Oord, A. v.~d.; Vinyals, O.; and Kavukcuoglu, K. 2017.
\newblock Neural discrete representation learning.
\newblock \emph{arXiv preprint arXiv:1711.00937}.

\bibitem[{Praneeth, brett koonce, and
  Ayinmehr(2019)}]{bedapudi_praneeth_2019_3584720}
Praneeth, B.; brett koonce; and Ayinmehr, A. 2019.
\newblock bedapudi6788/NudeNet: place for checkpoint files.

\bibitem[{Radford et~al.(2021)Radford, Kim, Hallacy, Ramesh, Goh, Agarwal,
  Sastry, Askell, Mishkin, Clark et~al.}]{radford2021learning}
Radford, A.; Kim, J.~W.; Hallacy, C.; Ramesh, A.; Goh, G.; Agarwal, S.; Sastry,
  G.; Askell, A.; Mishkin, P.; Clark, J.; et~al. 2021.
\newblock Learning transferable visual models from natural language
  supervision.
\newblock In \emph{International conference on machine learning}, 8748--8763.
  PMLR.

\bibitem[{Ramesh et~al.(2021)Ramesh, Pavlov, Goh, Gray, Voss, Radford, Chen,
  and Sutskever}]{ramesh2021zero}
Ramesh, A.; Pavlov, M.; Goh, G.; Gray, S.; Voss, C.; Radford, A.; Chen, M.; and
  Sutskever, I. 2021.
\newblock Zero-shot text-to-image generation.
\newblock In \emph{International Conference on Machine Learning}, 8821--8831.
  PMLR.

\bibitem[{Rando et~al.(2022)Rando, Paleka, Lindner, Heim, and
  Tram{\`e}r}]{rando2022red}
Rando, J.; Paleka, D.; Lindner, D.; Heim, L.; and Tram{\`e}r, F. 2022.
\newblock Red-teaming the stable diffusion safety filter.
\newblock \emph{arXiv preprint arXiv:2210.04610}.

\bibitem[{Razavi, Van~den Oord, and Vinyals(2019)}]{razavi2019generating}
Razavi, A.; Van~den Oord, A.; and Vinyals, O. 2019.
\newblock Generating diverse high-fidelity images with vq-vae-2.
\newblock \emph{Advances in neural information processing systems}, 32.

\bibitem[{Rombach et~al.(2022)Rombach, Blattmann, Lorenz, Esser, and
  Ommer}]{rombach2022high}
Rombach, R.; Blattmann, A.; Lorenz, D.; Esser, P.; and Ommer, B. 2022.
\newblock High-resolution image synthesis with latent diffusion models.
\newblock In \emph{Proceedings of the IEEE/CVF conference on computer vision
  and pattern recognition}, 10684--10695.

\bibitem[{Ruiz et~al.(2023)Ruiz, Li, Jampani, Pritch, Rubinstein, and
  Aberman}]{ruiz2023dreambooth}
Ruiz, N.; Li, Y.; Jampani, V.; Pritch, Y.; Rubinstein, M.; and Aberman, K.
  2023.
\newblock Dreambooth: Fine tuning text-to-image diffusion models for
  subject-driven generation.
\newblock In \emph{Proceedings of the IEEE/CVF Conference on Computer Vision
  and Pattern Recognition}, 22500--22510.

\bibitem[{Saharia et~al.(2022)Saharia, Chan, Saxena, Li, Whang, Denton,
  Ghasemipour, Gontijo~Lopes, Karagol~Ayan, Salimans
  et~al.}]{saharia2022photorealistic}
Saharia, C.; Chan, W.; Saxena, S.; Li, L.; Whang, J.; Denton, E.~L.;
  Ghasemipour, K.; Gontijo~Lopes, R.; Karagol~Ayan, B.; Salimans, T.; et~al.
  2022.
\newblock Photorealistic text-to-image diffusion models with deep language
  understanding.
\newblock \emph{Advances in Neural Information Processing Systems}, 35:
  36479--36494.

\bibitem[{Santurkar et~al.(2019)Santurkar, Ilyas, Tsipras, Engstrom, Tran, and
  Madry}]{santurkar2019image}
Santurkar, S.; Ilyas, A.; Tsipras, D.; Engstrom, L.; Tran, B.; and Madry, A.
  2019.
\newblock Image synthesis with a single (robust) classifier.
\newblock \emph{Advances in Neural Information Processing Systems}, 32.

\bibitem[{Schuhmann et~al.(2022)Schuhmann, Beaumont, Vencu, Gordon, Wightman,
  Cherti, Coombes, Katta, Mullis, Wortsman et~al.}]{schuhmann2022laion}
Schuhmann, C.; Beaumont, R.; Vencu, R.; Gordon, C.; Wightman, R.; Cherti, M.;
  Coombes, T.; Katta, A.; Mullis, C.; Wortsman, M.; et~al. 2022.
\newblock Laion-5b: An open large-scale dataset for training next generation
  image-text models.
\newblock \emph{Advances in Neural Information Processing Systems}, 35:
  25278--25294.

\bibitem[{Sohl-Dickstein et~al.(2015)Sohl-Dickstein, Weiss, Maheswaranathan,
  and Ganguli}]{sohl2015deep}
Sohl-Dickstein, J.; Weiss, E.; Maheswaranathan, N.; and Ganguli, S. 2015.
\newblock Deep unsupervised learning using nonequilibrium thermodynamics.
\newblock In \emph{International conference on machine learning}, 2256--2265.
  PMLR.

\bibitem[{Song, Meng, and Ermon(2020)}]{song2020denoising}
Song, J.; Meng, C.; and Ermon, S. 2020.
\newblock Denoising diffusion implicit models.
\newblock \emph{arXiv preprint arXiv:2010.02502}.

\bibitem[{Song et~al.(2020)Song, Sohl-Dickstein, Kingma, Kumar, Ermon, and
  Poole}]{song2020score}
Song, Y.; Sohl-Dickstein, J.; Kingma, D.~P.; Kumar, A.; Ermon, S.; and Poole,
  B. 2020.
\newblock Score-based generative modeling through stochastic differential
  equations.
\newblock \emph{arXiv preprint arXiv:2011.13456}.

\bibitem[{von Platen et~al.(2022)von Platen, Patil, Lozhkov, Cuenca, Lambert,
  Rasul, Davaadorj, and Wolf}]{von-platen-etal-2022-diffusers}
von Platen, P.; Patil, S.; Lozhkov, A.; Cuenca, P.; Lambert, N.; Rasul, K.;
  Davaadorj, M.; and Wolf, T. 2022.
\newblock Diffusers: State-of-the-art diffusion models.
\newblock \url{https://github.com/huggingface/diffusers}.

\bibitem[{Zhang et~al.(2023)Zhang, Wang, Xu, Wang, and Shi}]{zhang2023forget}
Zhang, E.; Wang, K.; Xu, X.; Wang, Z.; and Shi, H. 2023.
\newblock Forget-me-not: Learning to forget in text-to-image diffusion models.
\newblock \emph{arXiv preprint arXiv:2303.17591}.

\bibitem[{Zhang et~al.(2019)Zhang, Song, Gao, Chen, Bao, and
  Ma}]{zhang2019your}
Zhang, L.; Song, J.; Gao, A.; Chen, J.; Bao, C.; and Ma, K. 2019.
\newblock Be your own teacher: Improve the performance of convolutional neural
  networks via self distillation.
\newblock In \emph{Proceedings of the IEEE/CVF international conference on
  computer vision}, 3713--3722.

\end{thebibliography}
}
\clearpage
\begin{center}
\begin{LARGE}  
\textbf{Supplementary Materials\\}   
\end{LARGE}
\end{center}

\begin{figure}[h]
\centering
\includegraphics[width=3.2in, height=1.6in]{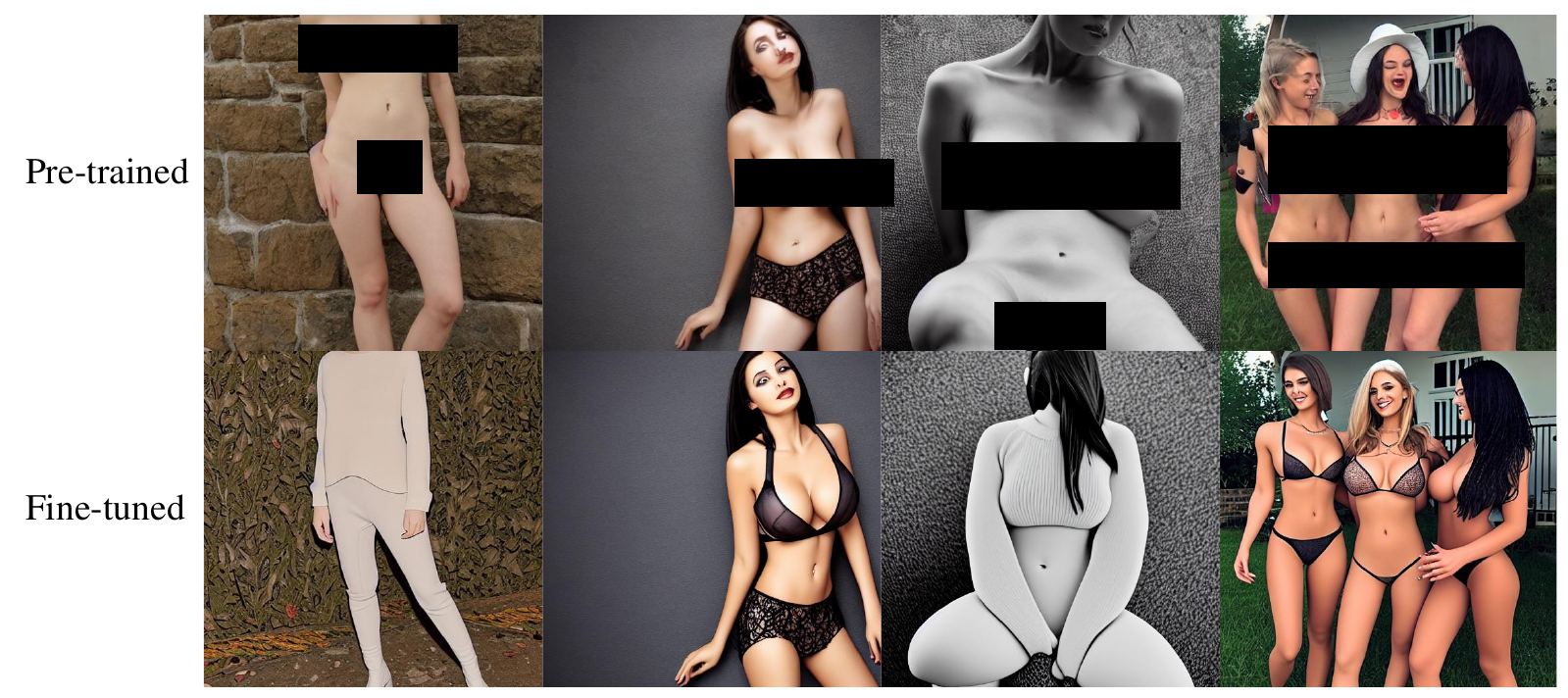}
\caption{Images before and after fine-tuning using Prompt-to-Prompt as our source for $\delta$ sampling.}
\label{fig:p2p}
\vspace{-8pt}
\end{figure}

\noindent\textbf{Implicit guidance with Prompt-to-Prompt.} Our method states that either explicit or implicit guidance $\delta$ works. We show its compatibility with implicit guidance with Prompt-to-Prompt \cite{hertz2022prompt}. Prompt-to-prompt allows re-weighting the attention maps while constraining their spatial distribution. If the attention maps corresponding to our target concept are suppressed, the learned prior of the model will introduce another concept to fill its place. Then, we use this modified $\epsilon$ to synthesize our guidance $\delta$. Although training with Prompt-to-Prompt presents its own instabilities (e.g. re-weighting hyperparameter, layer of application, number of timesteps to apply), we show in Fig \ref{fig:p2p}. that it produces good erased samples.

\noindent\textbf{Parameter Selection. } The current consensus is that text embeddings are contextualized in the cross-attention layers \cite{gandikota2023erasing,kumari2023ablating,kim2023towards,zhang2023forget}. Accordingly, previous approaches have chosen to update these layers as a standard practice. However, erasing fine-tuning relies on training with a very limited sample size, and the model is prone to overfitting. Furthermore, while the attention layer is of our interest, we wish to preserve the spatial priors learned by the key, query weights as much as possible. To this end, we completely bypass the weights responsible for synthesizing the attention maps and consider updating the final linear layer of the attention layer ($to\, out$) 
 \cite{von-platen-etal-2022-diffusers} of both cross and self-attention. This choice of parameter granted us more spatial consistency across all images. Additionally, the $to \, out$ layer processes the feature embedding yielded from every attention head. One can argue that the update on these shared weights has an implicit regulatory effect.

\begin{figure}[h]
\centering
\centerline{\includegraphics[width=1\columnwidth]{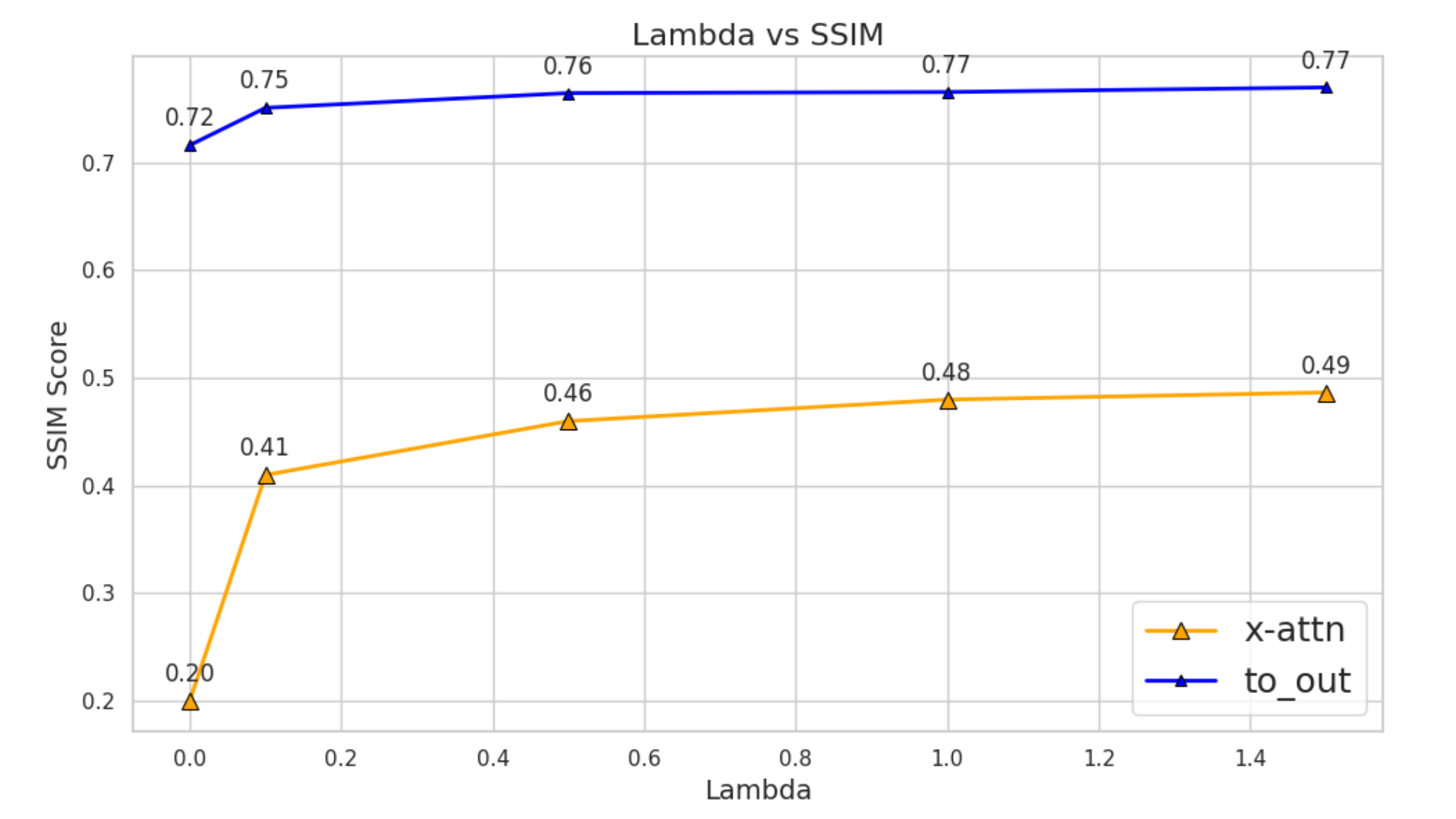}}
\caption{SSIM sensitivity for different $\lambda$ values. Optimizing the $to \, out$ layer regularizes the spatial priors to a greater degree.}
\label{fig:lambda2}
\vspace{-8pt}
\end{figure}

\noindent\textbf{Lambda Sensitivity. } In this section, we emphasize the role of $\lambda$ in our optimization. We state that $\lambda$ constrains the model from distribution shifts due to the erasing fine-tuning. Then, we can expect a higher spatial consistency after the optimization. In Fig. \ref{fig:lambda2}, we calculate the SSIM for 1,000 random objects compared to the object images generated by the base model SD v1.4. We show the sensitivity for models with cross-attention, and $to \, out$ layers updated. As soon as $\lambda > 1$, its spatial regularization effect is evident, fading away in a logarithmic fashion.

\begin{figure*}[ht]
 \centering
  \includegraphics[width=0.95\textwidth]{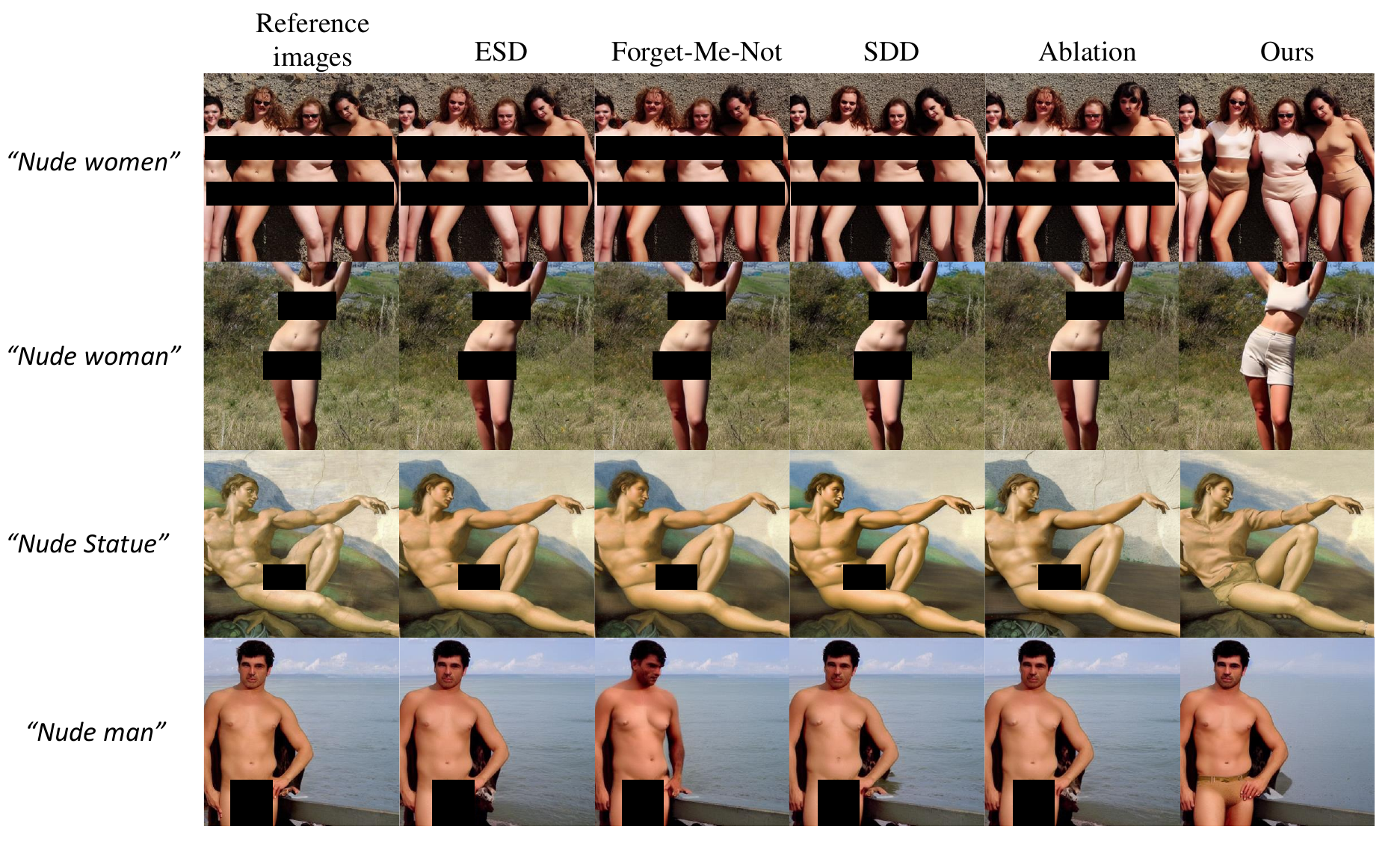}
\caption{DDIM inversion with respective text prompts. For other models, it does not purify even with the null text condition. }
 \label{fig:purification}
 \vspace{-8pt}
\end{figure*}

\noindent\textbf{Concept Purification.} A corollary from optimizing with our method is that our DDIM inversion \cite{dhariwal2021diffusion} conditioned on the erasing concept ``purifies" and erases the concept. To our surprise, when tested with the null or the erasing concept condition, none of the other baselines generated comparable results, shown in Fig. \ref{fig:purification} . This further shows how the underlying mechanism of our method is different from others. It is noticeable how the spatial consistency is near flawless. This ``purification'' capability may have applications to censorship preprocessing, such as erasing ``car plates'' to censor private information in the dataset.

\section{Additional Results and Details of Objective Derivation}
\begin{figure*}[h!]
 \centering
  \includegraphics[width=0.75\textwidth]{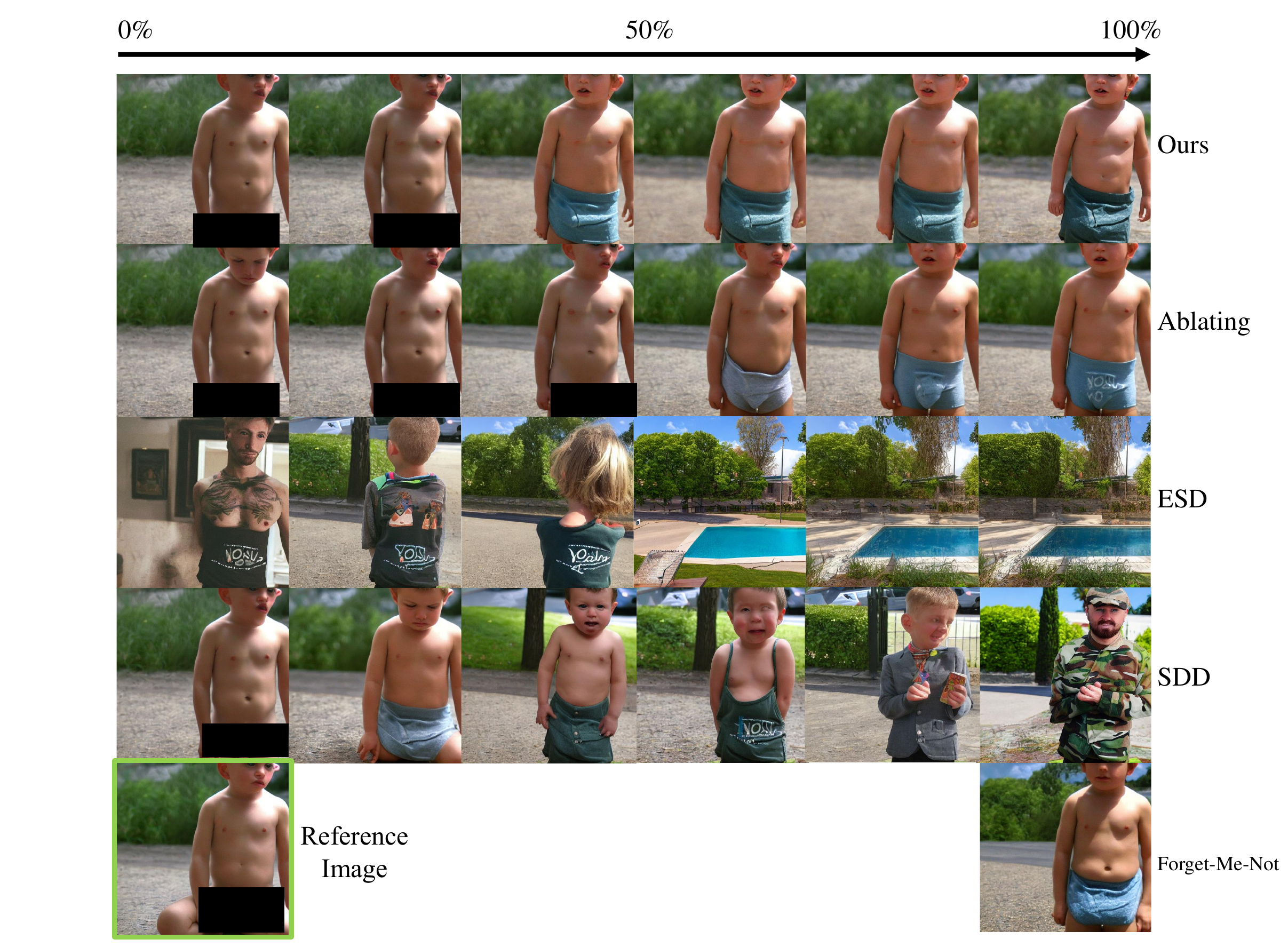}
\caption{Iteration timeline for images of the same seed prompted with "naked boy". Because the training ends at different rates, we normalized and expressed the progress in percentage. For the sake of demonstrating optimization convergence, we show iteration 400 in the last column of our model instead of our recommended iteration 200. Lastly, we set iteration 1,400 as the last step for SDD, as the model completely breaks at iteration 2,000. }
 \label{fig:nakedboy}
 \vspace{-8pt}
\end{figure*}

\begin{figure*}[h!]
 \centering
  \includegraphics[width=0.75\textwidth]{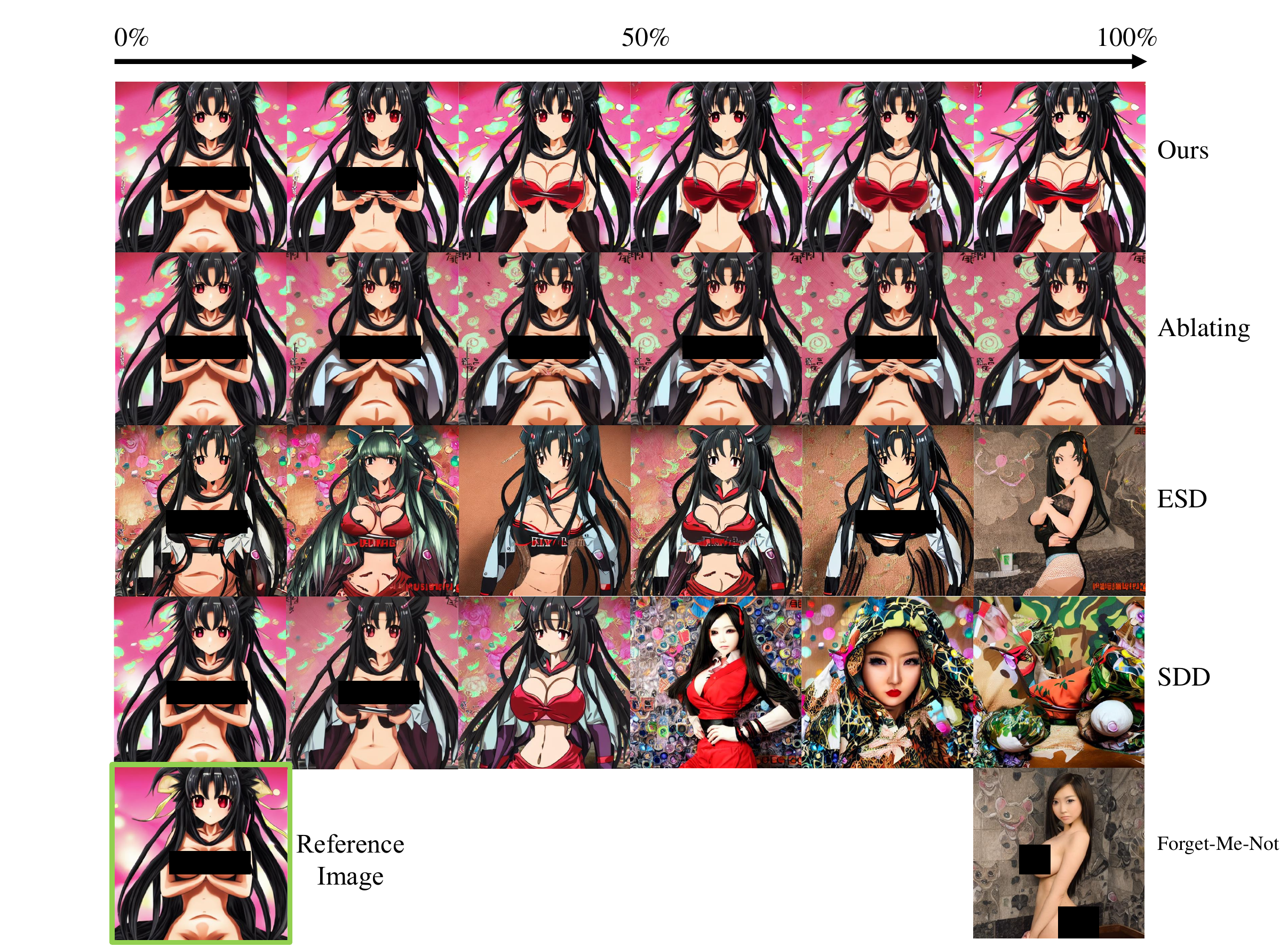}
\caption{Iteration timeline for images of the same seed prompted with ``Hentai". }
 \label{fig:hentai}
 \vspace{-8pt}
\end{figure*}

\begin{figure*}[h]
 \centering
  \includegraphics[width=0.73\textwidth]{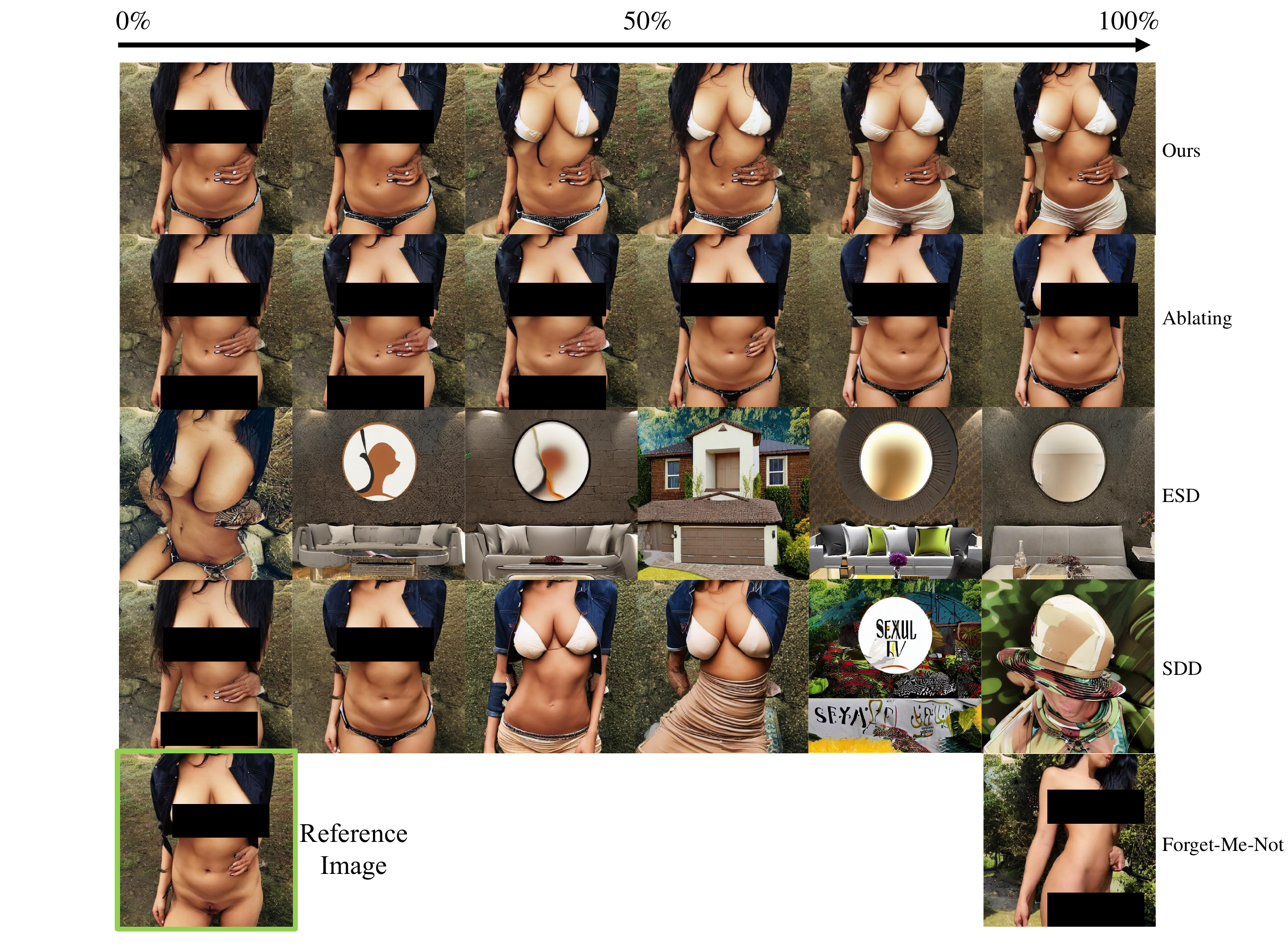}
\caption{Iteration timeline for images of the same seed prompted with ``sexual girl".}
 \label{fig:sexual_girl}
 \vspace{-8pt}
\end{figure*}

\noindent\textbf{Qualitative Results.} In this section, we provide qualitative results from every method. In Figs. \ref{fig:nakedboy},\ref{fig:hentai},\ref{fig:sexual_girl}, we show erasing of ``nudity'' from different baseline methods along with ours. Because the training ends at different rates, we normalized and expressed the progress in percentage. For the sake of demonstrating optimization convergence, we show iteration 400 in the last column of our model instead of our recommended iteration 200. Lastly, we set iteration 1,400 as the last step for SDD, as the model completely breaks at iteration 2,000.

\begin{figure*}[h!]
 \centering
  \includegraphics[width=0.7\textwidth]{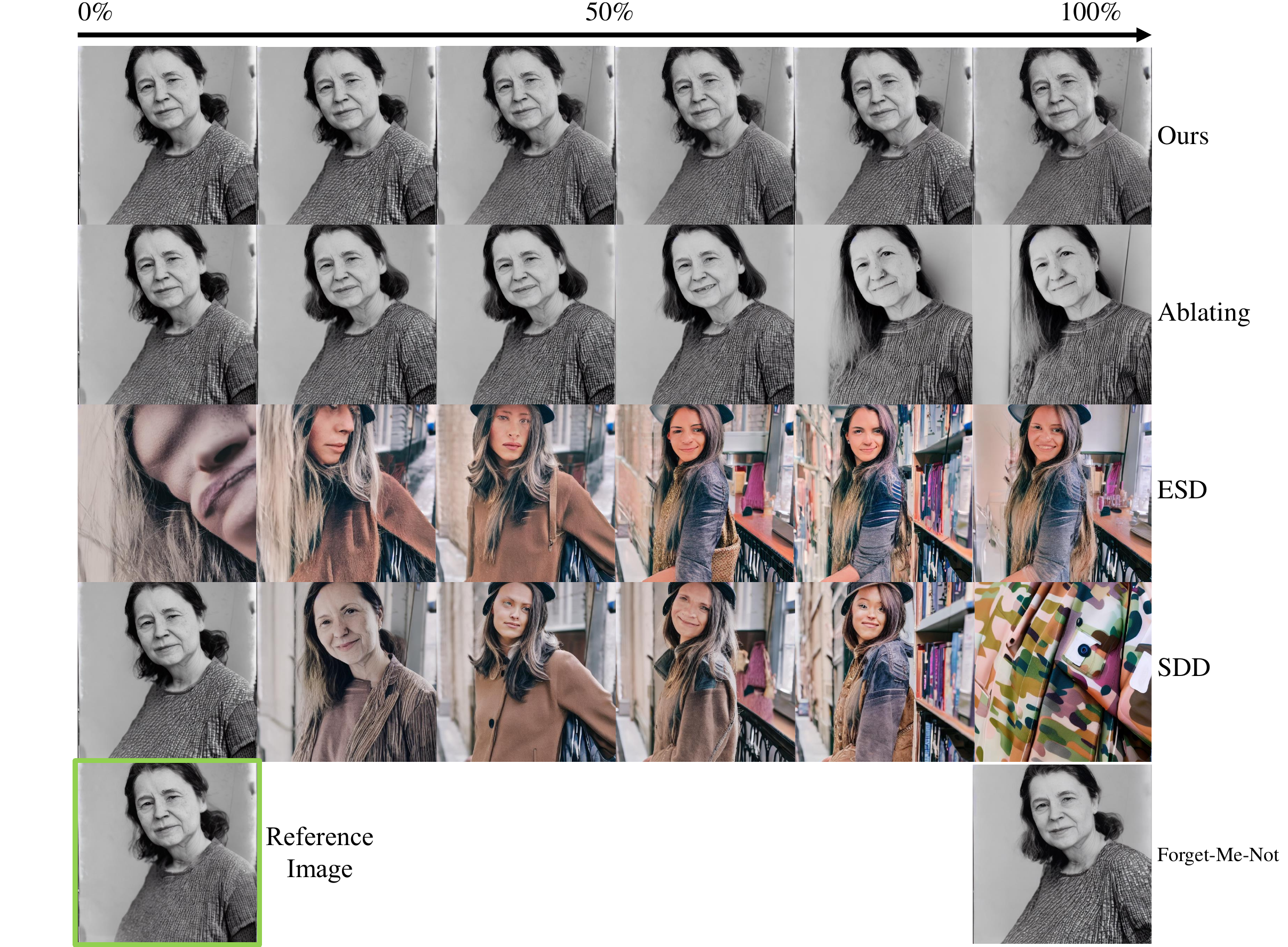}
\caption{Iteration timeline for images of the same seed prompted with ``photo of a woman". Here, we expect related concepts such as this to not change. Again, for our last column, we present the image for iteration 400 }
 \label{fig:woman}
 \vspace{-8pt}
\end{figure*}

Each shows a different paradigm of update. ESD shows fast but unstable erasing early on. Just as their objective function suggests, it is visible that all images with the concept ``nudity'' are mapped to near-unconditional images. SDD shows a more stable update. However, starting from iteration 1,000, the images collapse to the ``army" concept. While they introduced self-distillation to minimize undesirable oscillation in the supervision, this comes with a rather adverse outcome at the end of the training. While Forget-Me-Not shows great performance in erasing specific or memorized concepts, it fails to work with more general concepts such as ``nudity''. ``Ablating" shows good spatial consistency with consistency. However, it often fails to erase nudity. Lastly, our model shows that it erases and does not overfit or optimize more than necessary. When the training model learns the distribution of $\delta$, it converges and no more visual changes are attributed. Lastly, we present how concepts close to the erasing concept change over the iterations in Fig. \ref{fig:woman}

\noindent\textbf{Objective Formulation. }
Consider unconditional reverse noising process $P_{\theta}(z_{t}|z_{t+1})=\mathcal{N}(\mu,\Sigma)$, conditional reverse noising process is $P_{\theta}(z_{t}|z_{t+1},c)$. \cite{dhariwal2021diffusion} show that $P_{\theta}(z_{t}|z_{t+1},c) \thicksim \mathcal{N}(\mu+\gamma\Sigma g,\Sigma)$ where $g = \nabla_{z_t}\log(P_{\phi}(c|z_t))|_{z_t=\mu}$, $\gamma$ is guidance scale. Then, given some timestep $t$ and conditions $c, c'$, the distribution minimization between $P_{\theta^{\star}}(z_{t-1}|z_{t},c')$ and $P_{\theta}(z_{t-1}|z_{t},c))$ can be expressed as follows:
\begin{equation}
    \mathbb{E}_{z_t\sim P_{\theta^{\star}}(z_{t}|c')}[\mathbf{D_{KL}}(P_{\theta^{\star}}(z_{t-1}|z_{t},c')\|P_{\theta}(z_{t-1}|z_{t},c))]
\end{equation}
Here, the KL divergence can be formulated as: 
\begin{align}
    &\mathbf{D_{KL}}(P_{\theta^{\star}}(z_{t-1}|z_{t},c')\|P_{\theta}(z_{t-1}|z_{t},c))\\
    &=\mathbf{D_{KL}}(\mathcal{N}(\underbrace{\mu_{\theta^{\star}}+\gamma_1\Sigma g'}_{\tilde{\boldsymbol{\mu}}_{\theta^{\star}}},\Sigma)||\mathcal{N}(\underbrace{\mu_{\theta}+\gamma_2\Sigma g}_{\tilde{\boldsymbol{\mu}}_{\theta}},\Sigma))\\
    &=w'(t)||(\frac{1}{\sqrt{\alpha_t}}\boldsymbol{z}_t+\frac{1-\alpha_t}{\sqrt{\alpha_t}}\boldsymbol{s_{\theta^{\star}}}(\boldsymbol{z}_t,t,c'))\nonumber \\
    &\quad \quad \quad \quad \quad \quad-(\frac{1}{\sqrt{\alpha_t}}\boldsymbol{z}_t+\frac{1-\alpha_t}{\sqrt{\alpha_t}}\boldsymbol{s_{\theta}}(\boldsymbol{z}_t,t,c))||^2_2 \\
    &=w(t)||\boldsymbol{s_{\theta^{\star}}}(\boldsymbol{z}_t,t,c'))-\boldsymbol{s_{\theta}}(\boldsymbol{z}_t,t,c))||^2_2 \\
    &=w(t)||(\nabla_{z_t}\log(P_{\theta^{\star}}(z_t))+\gamma_1\nabla_{z_t}\log(P_{\theta^{\star}}(c'|z_t)))\nonumber \\
    &\quad \quad -(\nabla_{z_t}\log(P_{\theta}(z_t))+\gamma_2\nabla_{z_t}\log(P_{\theta}(c|z_t)))||\\
    &=w(t)||\underbrace{(\nabla_{z_t}\log(P_{\theta^{\star}}(z_t))-\nabla_{z_t}\log(P_{\theta}(z_t)))}_{\mathcal{L}_{U}\text{: unconditional loss term}} \nonumber \\
\end{align}
\begin{align}
    &-\underbrace{(\gamma_1\nabla_{z_t}\log(P_{\theta^{\star}}(c'|z_t))-\gamma_2\nabla_{z_t}\log(P_{\theta}(c|z_t)))}_{\mathcal{L}_{C}\text{: conditional loss term}}||^2_2\\
    &=w(t)||\mathcal{L}_{U}+\mathcal{L}_{C}||^2_2,
\end{align}
where $\gamma_1,\gamma_2$ are guidance scales, and $\mu_{\cdot}$ is the estimated denoising transition mean, $\tilde{\mu_{\theta}}$ is conditional guided denoising transition mean:
$$\boldsymbol{\tilde{\mu_{\theta}}}(\boldsymbol{z}_t,t,c)=\frac{1}{\sqrt{\alpha_t}}\boldsymbol{z}_t+\frac{1-\alpha_t}{\sqrt{\alpha_t}}\boldsymbol{s_{\theta}}(\boldsymbol{z}_t,t,c),$$
\noindent Here, $\alpha_t$ is scheduled noise variance, $\bar{\alpha}_t=\prod^{t}_{i=1}{\alpha_i}$, and $w(t),w'(t)$ are timestep-dependent loss weights as below: 
$$w(t)=\frac{2(1-\bar{\alpha_t})(1-\alpha_t)^2}{(1-\alpha_t)(1-\bar{\alpha_{t-1}})(\alpha_t)}$$
$$w'(t)=\frac{2(1-\bar{\alpha_t})}{(1-\alpha_t)(1-\bar{\alpha_{t-1}})}$$
\noindent Through our derivation, our initial KL divergence summarizes into:
\begin{equation*}
    \begin{aligned}
        &\mathbb{E}_{z_t\sim P_{\theta^{\star}}(z_{t}|c')}[\mathbf{D_{KL}}(P_{\theta^{\star}}(z_{t-1}|z_{t},c')\|P_{\theta}(z_{t-1}|z_{t},c))]\\
        &=\mathbb{E}_{z_t\sim P_{\theta^{\star}}(z_{t}|c')}[w(t)||\mathcal{L}_{U}+\mathcal{L}_{C}||^2_2]
    \end{aligned}
\end{equation*}
Then, by triangle inequality, $||\mathcal{L}_{U}+\mathcal{L}_{C}||_2$ gets $||\mathcal{L}_{U}||_2+||\mathcal{L}_{C}||_2$ as upper bound. 
Therefore, minimizing our loss $||\mathcal{L}_{U}||_2+||\mathcal{L}_{C}||_2$ is equivalent to minimizing the upper bound of $||\mathcal{L}_{U}+\mathcal{L}_{C}||_2$\\


\

\noindent \textbf{Experiment Setting. }
To sample $\delta$, the following are the hyperparameters for erasing "nudity": guidance  scale: $\gamma= 7.5$, lambda : $\lambda= 1$, max sampling step: $T=35$, and training parameter set is either "cross attention" or "$to \, out$" layer. For $c'$, we choose the following concepts:

\begin{equation}
\begin{aligned}
&\{c_{\text{text}}= \text{{sexual}}, g_c= -7.5, t_{c_{\text{high}}}=\lfloor 0.35T \rfloor,t_{c_{\text{low}}}=\lfloor T \rfloor,\kappa=0.95   \} \\
&\{c_{\text{text}}= \text{{bikini}}, g_c= 6.5, t_{c_{\text{high}}}=\lfloor 0.35T \rfloor,t_{c_{\text{low}}}=\lfloor T \rfloor,\kappa=0.95   \}\\
&\{c_{\text{text}}= \text{{pants}}, g_c= 6.5, t_{c_{\text{high}}}=\lfloor 0.35T \rfloor,t_{c_{\text{low}}}=\lfloor T \rfloor,\kappa=0.95   \}
\end{aligned}
\end{equation}

For Concept Purification ("nudity"), guidance  scale: $\gamma= 7.5$, lambda : $\lambda= 0$, max sampling step: $T=35$, training parameter set: ``cross attention". For $c'$, we choose the following concepts:

\begin{equation}
\begin{aligned}
&\{c_{\text{text}}= \text{{sexual}}, g_c= -7.5, t_{c_{\text{high}}}=\lfloor 0.35T \rfloor,t_{c_{\text{low}}}=\lfloor T \rfloor,\kappa=0.95   \}\\
&\{c_{\text{text}}= \text{{bikini}}, g_c= 6.5, t_{c_{\text{high}}}=\lfloor 0.35T \rfloor,t_{c_{\text{low}}}=\lfloor T \rfloor,\kappa=0.95   \}\\
&\{c_{\text{text}}= \text{{pants}}, g_c= 6.5, t_{c_{\text{high}}}=\lfloor 0.35T \rfloor,t_{c_{\text{low}}}=\lfloor T \rfloor,\kappa=0.95   \}
\end{aligned}
\end{equation}

\end{document}